\documentclass[journal]{IEEEtran}
\usepackage{amsmath, amsfonts}
\usepackage{orcidlink}

\usepackage[nocompress]{cite}
\hypersetup{
colorlinks,
citecolor=blue,
linkcolor=blue,
urlcolor=blue
}

\usepackage{mathrsfs}
\usepackage{algorithmic}
\usepackage{amssymb}
\usepackage{algorithm}
\usepackage{array}
\usepackage[caption=false, font=normalsize, labelfont=rm, textfont=rm]{subfig}
\usepackage{textcomp}
\usepackage{stfloats}
\usepackage{url}
\usepackage{bm}
\usepackage{verbatim}
\usepackage{graphicx}
\usepackage{siunitx}
\usepackage{graphicx}
\usepackage[normalem]{ulem}
\usepackage{array}
\usepackage{makecell}
\usepackage{multirow}
\usepackage{booktabs}

\begin{document}

\title{Structural-Spectral Graph Convolution with Evidential Edge Learning for Hyperspectral Image Clustering}

\author{
  Jianhan~Qi$^{\orcidlink{0009-0004-3810-083X}}$,
  Yuheng~Jia$^{\orcidlink{0000-0002-3907-6550}}$,~\IEEEmembership{Member,~IEEE,}
  Hui~Liu$^{\orcidlink{0000-0003-2159-025X}}$,
  Junhui~Hou$^{\orcidlink{0000-0003-3431-2021}}$,~\IEEEmembership{Senior Member,~IEEE}
  \thanks{This work was supported by the National Natural Science Foundation of China (Grants U24A20322, 62576094, and 62422118), the Hong Kong UGC (Grants UGC/FDS11/E03/24 and UGC/FDS11/E03/25), and the Hong Kong Research Grants Council (Grant 11219324). This work was also supported by the Big Data Computing Center of Southeast University. \textit{(Corresponding author: Yuheng Jia.)}}
  \thanks{Jianhan Qi is with the School of Software Engineering, Southeast University, Nanjing 211189, China (e-mail: qjh@seu.edu.cn).}
  \thanks{Yuheng Jia is with the School of Computer Science and Engineering, Southeast University, Nanjing 211189, China, and with the Key Laboratory of New Generation Artificial Intelligence Technology and Its Interdisciplinary Applications (Southeast University), Ministry of Education, Nanjing 211189, China, and also with the Department of Computing and Information Sciences, Saint Francis University, Hong Kong, China (e-mail: yhjia@seu.edu.cn).}
  \thanks{Hui Liu is with the Department of Computing and Information Sciences, Saint Francis University, Hong Kong, China (e-mail: h2liu@sfu.edu.hk).}
  \thanks{Junhui Hou is with the Department of Computer Science, City University of Hong Kong, Hong Kong, China (e-mail: jh.hou@cityu.edu.hk).}
}

\markboth{Manuscript Submitted to IEEE TIP}%
{Shell \MakeLowercase{\textit{et al.}}: A Sample Article Using IEEEtran.cls for IEEE Journals}


\maketitle

\begin{abstract}
Hyperspectral image (HSI) clustering groups pixels into clusters without labeled data, which is an important yet challenging task. For large-scale HSIs, most methods rely on superpixel segmentation and perform superpixel-level clustering based on graph neural networks (GNNs). However, existing GNNs cannot fully exploit the spectral information of the input HSI, and the inaccurate superpixel topological graph may lead to the confusion of different class semantics during information aggregation. To address these challenges, we first propose a structural-spectral graph convolutional operator (SSGCO) tailored for graph-structured HSI superpixels to improve their representation quality through the co-extraction of spatial and spectral features. Second, we propose an evidence-guided adaptive edge learning (EGAEL) module that adaptively predicts and refines edge weights in the superpixel topological graph. We integrate the proposed method into a contrastive learning framework to achieve clustering, where representation learning and clustering are simultaneously conducted. Experiments demonstrate that the proposed method improves clustering accuracy by 2.61\%, 6.06\%, 4.96\% and 3.15\% over the best compared methods on four HSI datasets. Our code is available at \url{https://github.com/jhqi/SSGCO-EGAEL}.
\end{abstract}

\begin{IEEEkeywords}
Hyperspectral image, clustering, superpixel, graph neural networks, contrastive learning.
\end{IEEEkeywords}

\section{Introduction\label{introduction}}
\IEEEPARstart{H}{yperspectral} image (HSI) captures the reflectance of land covers across hundreds of contiguous narrow bands, providing rich and precise spectral information. Classifying the pixels of HSI into different categories is a fundamental task \cite{hsi_cls_1,hsi_cls_2,hsi_cls_3,hsi_cls_4,hsi_cls_5,hsi_cls_6,hsi_cls_7}, as it describes the spatial distribution of different land covers. This technique is widely applied in various fields, such as precision agriculture \cite{app_precision_agriculture}, environmental monitoring \cite{app_environmental_monitoring}, mineral exploration \cite{app_mineral_exploration}, and urban planning \cite{app_urban_planning}.

\begin{figure}[ht]
  \centering
  \hfill
  \subfloat[]{\includegraphics[width=0.48\linewidth]{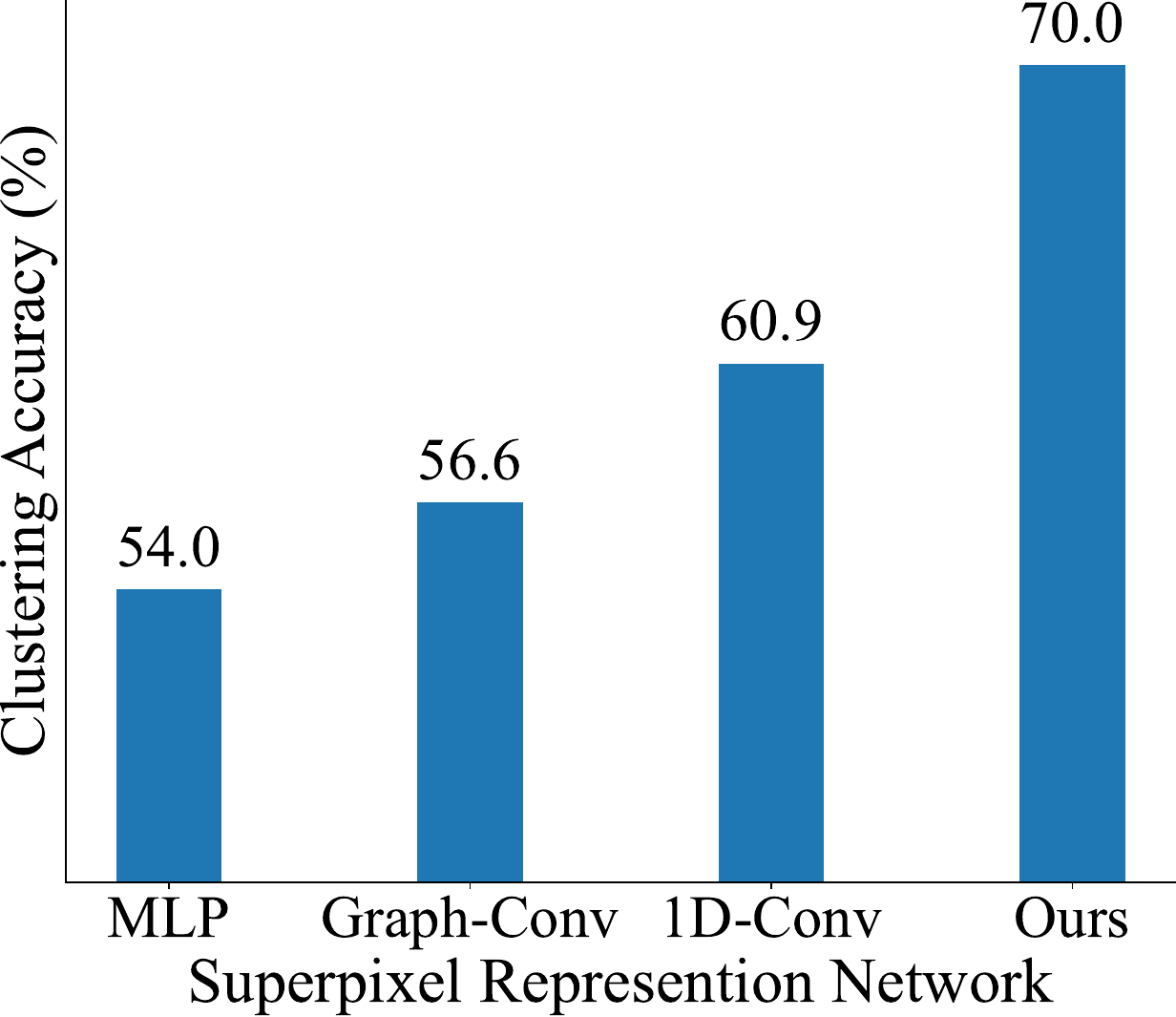}\label{pu_diff_net}}
  \hfill
  \subfloat[]{\includegraphics[width=0.48\linewidth]{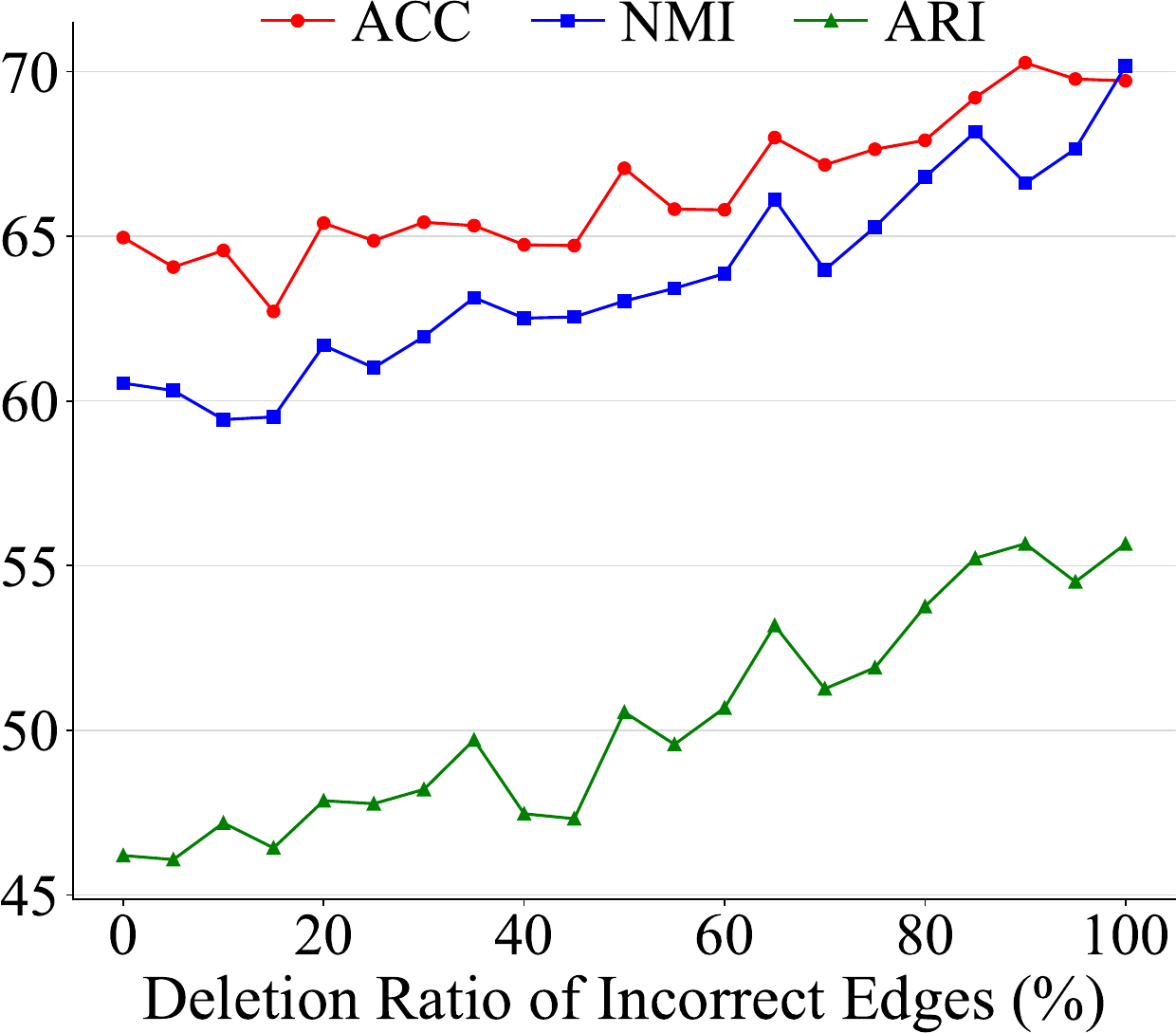}\label{pu_deletion_ratio}}
  \hfill
  \caption{Motivation of the proposed method: (a) clustering accuracy of different superpixel feature representation networks on \textit{Pavia University} dataset, (b) the impact of deleting incorrect edges on \textit{Pavia University} dataset.}
  \label{motivation_fig}
\end{figure}

Due to the high cost of obtaining HSI pixel annotations, unsupervised clustering methods that explore the relationships between pixels without labels have gained increasing attention. Traditional clustering methods, e.g., K-means \cite{kmeans}, fuzzy c-means \cite{fuzzycmeans} and subspace clustering \cite{ssc}, have been applied to HSI clustering, but with relatively poor clustering performance. This is because those methods usually perform clustering based on the raw feature of HSI, which is susceptible to spectral noise of HSI caused by the illumination condition or the viewing angle \cite{spectral_noise}. To this end, research has increasingly turned to deep clustering methods \cite{dec,pscpc,s2gcl}, which leverage deep neural networks to learn discriminative feature representations, achieving effective clustering results under end-to-end training. However, the massive number of pixels in HSI may lead to high computational cost. Consequently, many studies employed superpixel segmentation techniques \cite{ers, slic, lsc} and transformed the clustering task from the pixel level to the superpixel level to reduce the computational cost. Superpixel-level clustering has currently become the mainstream approach for large-scale HSI clustering \cite{superpixel_clustering_1,superpixel_clustering_2,superpixel_clustering_3,sglsc,ncsc}.

Given the irregular shapes and clear spatial distribution, superpixels are naturally modeled as graph-structured data, where each superpixel acts as a node and the topological graph is constructed based on their spatial adjacency or spectral similarity. With the powerful capability of information aggregation, graph neural networks (GNNs) have excelled in graph learning tasks \cite{gcn,gat}. Consequently, several studies have employed GNNs for superpixel representation learning to incorporate topological information into superpixel embeddings and then conducted GNN-based clustering, which can be mainly categorized into two types. The first type is based on graph autoencoders (GAEs) \cite{gae}, which typically reconstruct the graph adjacency matrix from node embeddings, and introduce an auxiliary distribution loss \cite{dec} for self-supervised training \cite{lowpass,EGRC,sdst,dgae,graph_clustering}. The second type utilizes contrastive clustering frameworks \cite{moco, byol, propos}, conducting alignment or contrast between samples or prototypes to improve the discriminability of superpixel representations \cite{pscpc,s2gcl,spgcc}. Although the above methods have achieved promising results, \textbf{two critical limitations still exist}.

The first limitation concerns insufficient feature extraction from superpixels. Besides the graph structure containing rich spatial information, the spectra of superpixels exhibit continuity and redundancy \cite{survey}. While existing GNNs can effectively aggregate spatial information within the neighborhood, the projective transformations (i.e., linear layers) in their networks struggle to capture complex patterns of HSI spectra. Conversely, convolutional neural networks (CNNs) are adept at extracting high-order spectral features, but fail to exploit the superpixel topological information. To illustrate representation learning abilities of different networks, in Fig. \ref{pu_diff_net}, we conducted a comparison among MLP, 1D convolution (1D-Conv), graph convolution (Graph-Conv) and our proposed operator designed to extract both spatial and spectral features, indicating that comprehensive superpixel feature extraction significantly enhances the clustering accuracy.

The second limitation is the inaccurate superpixel topological graph. Typically, the superpixel topological graph is constructed based on either spatial adjacency or spectral similarity. The former connects spatially adjacent superpixels as neighbors \cite{sglsc, spgcc}, while the latter determines neighbors based on spectral similarity \cite{lowpass,sdst,dgae}. EGRC-Net\cite{EGRC} adaptively improves the initial graph by adding edges between the nearest node pairs based on nodes' current embeddings. However, it fails to take into account the reliability of existing edges. Since the superpixel topological graph is constructed without supervision, both spatial and spectral graphs inevitably contain incorrect edges that connect two superpixel nodes of different classes. Accordingly, these incorrect edges bring semantically different neighbors, leading to distorted node representations and confused class semantics during GNN aggregation. Fig. \ref{pu_deletion_ratio} illustrates the impact of deleting incorrect edges on clustering results. As the deletion ratio increases, clustering performance gradually improves, highlighting the importance of dealing with incorrect edges in the graph. 

Based on the above analyses, we propose a structural-spectral graph convolutional operator (SSGCO) to effectively extract spatial and spectral features for HSI superpixels, and an evidence-guided adaptive edge learning (EGAEL) module to refine the superpixel topological graph. SSGCO performs 1D convolution along the spectral dimension to capture high-order spectral patterns, while simultaneously performing aggregation within the neighborhood to exploit spatial information, which is tailored to HSI superpixel representation learning. Fig. \ref{pu_diff_net} shows that the operator itself yields a notable improvement of 9.1\% in clustering accuracy. For EGAEL, we represent each edge by the embeddings of its two endpoint nodes and re-predict the edge weight through a learnable MLP adaptively during training, thereby reducing the weights of incorrect edges. Besides, we propose to use the empirical edge weight computed based on clustering confidence and node similarity as evidence to guide the edge weight prediction. The superpixel adjacency matrix is then momentum updated with predicted edge weights. Finally, we integrate the proposed SSGCO and EGAEL into a contrastive clustering framework, where pixels randomly sampled from superpixels act as augmented views. Neighborhood alignment and prototype contrast are combined as clustering-oriented training objectives, facilitating a mutual promotion between superpixel representation learning and clustering.

The main contributions of this paper are summarized as follows.
\begin{enumerate}
\item[1)] We propose a novel structural-spectral graph convolutional operator, specifically designed for HSI superpixels, which achieves comprehensive co-extraction of both deep spectral patterns and spatial information.

\item[2)] We propose an evidence-guided adaptive edge learning module to reduce the weights of incorrect edges, mitigating the impact of the inaccurate superpixel topological graph.

\item[3)] We integrate our method into the contrastive clustering framework and achieve promising clustering performance. Extensive experiments on four HSI benchmark datasets demonstrate the superiority of our method. In particular, our method achieves an average improvement of 4.20\% in clustering accuracy on the four datasets.
\end{enumerate}

\begin{figure*}[ht]
  \centering
  \includegraphics[width=\textwidth]{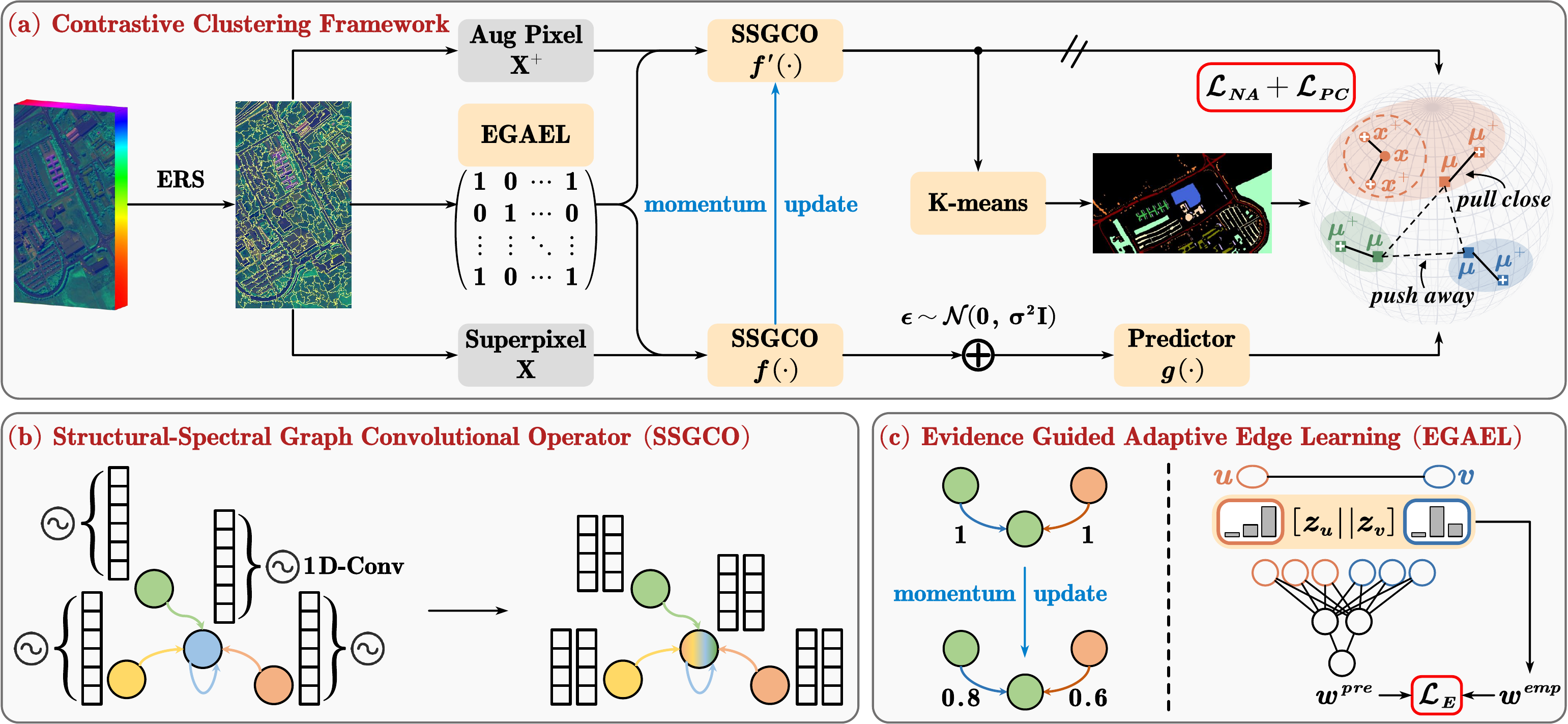}
  \caption{The diagram of the proposed method: \textbf{(a)} the contrastive clustering framework consists of an online network $f(\cdot)$ and a target network $f'(\cdot)$, which take superpixels and augmented views as inputs. Neighborhood alignment (NA) and prototype contrast (PC) are adopted as training objectives; \textbf{(b)} Structural-spectral graph convolutional operator (SSGCO) sequentially performs 1D convolution (1D-Conv) and graph convolution in each layer to extract both spatial and spectral features of superpixels; \textbf{(c)} In evidence guided adaptive edge learning (EGAEL), each edge is represented by concatenating the soft assignments of its two endpoint nodes, and input into an MLP to get the predicted edge weight $w^{pre}$. The empirical edge weight $w^{emp}$ computed with node clustering confidence and similarity is adopted as guidance. The adjacency matrix is momentum updated based on $w^{pre}$.}
  \label{pipeline}
\end{figure*}

\section{Data Preprocessing\label{data_preprocessing}}
To reduce the computational cost on large HSI, we employ entropy rate superpixel segmentation (ERS) \cite{ers} to segment the input HSI $\mathcal{H}$ into $M$ non-overlapped superpixels, where each pixel belongs to exactly one superpixel. Specifically, we perform principal components analysis (PCA) on pixel features and then retain the first principal component to obtain the gray-scale image as the input of ERS.

Then, we apply PCA on raw pixel features again, and retain only $d$ components to further reduce the data dimensionality \cite{dgae}. For the $i$-th pixel $p_i$ and the $j$-th superpixel $sp_j$, their feature vectors are denoted as $\boldsymbol{x}^{p}_{i} \in \mathbb{R}^{d}$ and $\boldsymbol{x}_{j} \in \mathbb{R}^{d}$, following:
\begin{equation}
  \label{sp_feat}
  \boldsymbol{x}_{j} = \frac{1}{\vert sp_{j} \vert}\sum_{p_i \in sp_j}{\boldsymbol{x}^{p}_{i}},
\end{equation}
where $\vert sp_{j} \vert$ denotes the number of pixels in $sp_j$. Thus, features vectors of all the superpixels are represented as $\mathbf{X} = [\boldsymbol{x}_{1};...;\boldsymbol{x}_{M}]\in \mathbb{R}^{M \times d}$, where each row corresponds to a superpixel.

Next, we build the superpixel topological graph based on the spatial adjacency of superpixels. Given two pixels $p_{m}$ and $p_{n}$, where $p_{m}$ belongs to $sp_{i}$ and $p_{n}$ belongs to $sp_{j}$, if $p_{m}$ and $p_{n}$ are adjacent (i.e., the Manhattan distance between their coordinates is equal to 1), then $sp_{i}$ and $sp_{j}$ are also adjacent. The superpixel adjacency matrix $\mathbf{A}\in \mathbb{R}^{M \times M}$ is defined as:
\begin{equation}
\label{adj}
\mathbf{A}_{i,j}= \mathbf{A}_{j,i}=
\begin{cases}
  1, & \exists p_{m} \in sp_{i}, p_{n} \in sp_{j}, \mathrm{dis}(p_{m}, p_{n})=1,\\
  0, & \mathrm{else},
\end{cases}
\end{equation}
where $\mathrm{dis}(\cdot)$ denotes the Manhattan distance. Following \cite{gcn}, we add self-loop to $\mathbf{A}$ and employ the re-normalization trick:
\begin{equation}
  \label{self_loop}\widetilde{\mathbf{A}}=\mathbf{A}+\mathbf{I}_{M},
\end{equation}
\begin{equation}
  \label{degree}\widetilde{\mathbf{D}}_{i,i}= \sum\nolimits_{j}{\widetilde{\mathbf{A}}_{i,j}}
\end{equation}
\begin{equation}
\label{renorm}
\widehat{\mathbf{A}}= \widetilde{\mathbf{D}}^{-\frac{1}{2}}\widetilde{\mathbf{A}}\widetilde{\mathbf{D}}^{-\frac{1}{2}},
\end{equation}
where $\mathbf{I}_{M} \in \mathbb{R}^{M \times M}$ is the identity matrix, $\widetilde{\mathbf{D}}$ is the degree matrix with self-loop, and $\widehat{\mathbf{A}}$ is the symmetrically normalized adjacency matrix.

\section{Proposed Method\label{proposed_method}}
In this section, we first introduce the adopted contrastive clustering framework, which leverages neighborhood alignment and prototype contrast as training objectives. Second, we introduce the proposed structural-spectral graph convolutional operator, which aims to extract both spatial and spectral features from superpixels. Third, we present the proposed evidence-guided adaptive edge learning to mitigate the impact of incorrect edges in the superpixel topological graph. Finally, we introduce the training procedure in detail. The overall pipeline of our method is illustrated in Fig. \ref{pipeline}.

\subsection{Contrastive Clustering Framework\label{ccf}}
We achieve HSI superpixel clustering under a contrastive clustering framework. Specifically, following \cite{byol, propos}, we adopt an online network $f(\cdot)$ and a target network $f'(\cdot)$ with identical architecture but unshared parameters as superpixel encoders. To break the symmetry between $f(\cdot)$ and $f'(\cdot)$, and prevent representation collapse, predictor $g(\cdot)$ consisting of two-layer MLP is introduced after $f(\cdot)$. Neighborhood alignment and prototype contrast are combined as clustering-oriented training objectives.

For neighborhood alignment, we first construct reliable positive samples for superpixels through data augmentation. Given superpixel $\boldsymbol{x}_i$, we randomly select one internal pixel as its augmented view $\boldsymbol{x}_i^{+}$ \cite{spgcc}. Then, based on the assumption that representations within a local neighborhood in the embedding space share similar semantics, superpixel representations are encouraged to be similar to not only their augmented views but also neighboring samples, thus improving intra-cluster compactness. Instead of directly sampling from the neighborhood, adopting the reparameterization trick ensures proper gradient back-propagation. For superpixel $\boldsymbol{x}_i$, its neighborhood in the embedding space can be represented as:
\begin{equation}
  \label{neighboring}
  \boldsymbol{v}_i= f(\boldsymbol{x}_i)+\sigma\boldsymbol{\epsilon},
\end{equation}
where $\boldsymbol{\epsilon} \sim \mathcal{N}(0, \mathbf{I})$ and $\sigma$ is a hyperparameter that controls the margin of the neighborhood. By minimizing the distance between samples and corresponding augmented views, the neighborhood alignment is formally defined as:
\begin{equation}
  \label{na}
  \begin{split}
    \mathcal{L}_{NA} & = \frac{1}{M} \sum_{i=1}^{M} \Vert g(\boldsymbol{v}_i)-f'(\boldsymbol{x}_i^{+})\Vert_{2}^{2} \\
    & = \frac{1}{M} \sum_{i=1}^{M} \Vert g(f(\boldsymbol{x}_i)+\sigma\boldsymbol{\epsilon})-f'(\boldsymbol{x}_i^{+})\Vert_{2}^{2}.
  \end{split}
\end{equation}

For prototype contrast, we first perform $l_2$-normalization on $f'(\boldsymbol{x})$ and conduct spherical K-means to obtain a clustering assignment. Let $K$ denote the total number of classes and $S_k$ denotes the set of samples assigned to the $k$-th cluster, then two views of the $k$-th prototype, $\boldsymbol{\mu}_k$ and $\boldsymbol{\mu}_k^+$, are defined as the average of sample embeddings in the cluster:
\begin{equation}
  \label{prototype}
  \boldsymbol{\mu}_k = \frac{\sum_{x \in S_k} f(\boldsymbol{x})}{\Vert \sum_{x \in S_k} f(\boldsymbol{x}) \Vert_2} , \quad \boldsymbol{\mu}_k^+ = \frac{\sum_{x \in S_k} f'(\boldsymbol{x}^+)}{\Vert \sum_{x \in S_k} f'(\boldsymbol{x}^+) \Vert_2}.
\end{equation}
Given one prototype, the remaining $K-1$ prototypes are definitely negative examples. We construct a contrastive loss \cite{infonce} to encourage the alignment between different views of the same prototype, and increase the inter-cluster distance, i.e.,
\begin{equation}
  \label{pc}
  \mathcal{L}_{PC}=\frac{1}{K}\sum_{k=1}^{K}-\log\frac{\exp(\boldsymbol{\mu}_k \cdot \boldsymbol{\mu}_k^+/\tau)}{\sum_{j=1}^K\exp(\boldsymbol{\mu}_k \cdot \boldsymbol{\mu}_j^+/\tau)},
\end{equation}
where $\tau \in (0,1)$ is the temperature scaling. With the weight hyperparameter $\alpha > 0$, the total loss for contrastive clustering is defined as:
\begin{equation}
  \label{cc_loss}
  \mathcal{L}_{CC} = \mathcal{L}_{NA} + \alpha \mathcal{L}_{PC}.
\end{equation}

For optimization, $f(\cdot)$ with parameters $\boldsymbol{\theta}$ is updated through gradient back-propagation, while $f'(\cdot)$ with parameters $\boldsymbol{\theta}'$ is momentum updated, following:
\begin{equation}
  \label{target_mom_update}
  \boldsymbol{\theta}' = m\boldsymbol{\theta}' + (1-m)\boldsymbol{\theta},
\end{equation}
where $m \in (0, 1)$ is the momentum coefficient.

\subsection{Structural-Spectral Graph Convolutional Operator}
While existing GNNs excel at leveraging spatial information of HSI based on the superpixel topological graph, they often inadequately extract deep spectral features. Their reliance on projective transformations (i.e., linear layers) limits the capability to deal with the inherent continuity and redundancy of HSI spectra. To address this limitation, we propose a novel structural-spectral graph convolutional operator (SSGCO) to enhance the superpixel representation learning. SSGCO enriches the information aggregation capability of GNNs by infusing dedicated 1D convolutional kernels that target high-order spectral feature extraction.

Based on the standard multi-layer GCN, each SSGCO layer sequentially applies 1D convolution to extract deep spectral patterns, followed by graph convolution to aggregate spatial information, as illustrated in Fig. \ref{pipeline}b. SSGCO is formally defined as:
\begin{equation}
\label{ssgco_eq}
\mathbf{H}_{(l)}=\phi(\widehat{\mathbf{A}} (\mathbf{H}_{(l-1)} \circledast_{\scriptscriptstyle \mathrm{1D}} \mathbf{W}^{c}_{(l)}) \mathbf{W}^{g}_{(l)}),
\end{equation}
where $\mathbf{H}_{(l)}$ is the output of the $l$-th layer and $\mathbf{H}_{(0)}=\mathbf{X}$, $\circledast_{\scriptscriptstyle \mathrm{1D}}$ denotes 1D convolutional operator, $\mathbf{W}^{c}_{(l)}$ and $\mathbf{W}^{g}_{(l)}$ are trainable parameters in the $l$-th layer corresponding to 1D convolution and graph convolution, $\phi(\cdot)$ is non-linear activation function. Moreover, we conduct batch normalization (BN) \cite{bn} after each layer to stabilize the training.

We show a specific example of the proposed SSGCO in Table \ref{ssgco_forward_propagation}, where the input feature dimension $d$ equals to 20 and SSGCO consists of two layers. For 1D convolution (1D-Conv), kernel sizes are set to 7 and 5, numbers of output channels are set to 8 and 16, paddings and strides are set to 0 and 1. For graph convolution (Graph-Conv), we reshape both the input and output representations to keep the dimension alignment. Notably, the kernels of 1D-Conv and Graph-Conv are denoted as (\#out channel, \#kernel size) and (\#in dim, \#out dim), respectively.
\begin{table}[t]
  \centering
  \caption{Structural-Spectral Graph Convolutional Operator Forward-Propagation.\label{ssgco_forward_propagation}}
  \renewcommand\arraystretch{1.1}
  \begin{tabular}{ccc} 
  \toprule
  Layer      & Kernel     & Output Dimension  \\ 
  \midrule
  Input      & -          & ($M$, 20)        \\
  Reshape    & -          & ($M$, 1, 20)       \\
  1D-Conv    & (8, 7)     & ($M$, 8, 14)       \\
  BN         & -          & ($M$, 8, 14)       \\
  Reshape    & -          & ($M$, 112)         \\
  Graph-Conv & (112, 112) & ($M$, 112)     \\
  BN         & -          & ($M$, 112)      \\
  Reshape    & -          & ($M$, 8, 14)       \\
  1D-Conv    & (16, 5)    & ($M$, 16, 10)    \\
  BN         & -          & ($M$, 16, 10)      \\
  Reshape    & -          & ($M$, 160)         \\
  Graph-Conv & (160, 160) & ($M$, 160)       \\
  BN         & -          & ($M$, 160)        \\
  \bottomrule
  \end{tabular}
\end{table}

\subsection{Evidence-guided Adaptive Edge Learning}
The superpixel topological graph is usually constructed based on the spatial adjacency. However, incorrect edges that connect superpixels of different classes are harmful to information aggregation in GNNs as they may confuse the semantics of different land covers. To this end, we propose an evidence-guided adaptive edge learning (EGAEL) module to reduce the weights of incorrect edges and improve the quality of the adjacency matrix.

First, we represent each edge by its two endpoint superpixel nodes. Specifically, given an edge $(u, v) \in \mathcal{E}$ where $\mathcal{E}$ denotes the set of all edges in the initial graph, nodes $u$ and $v$ are represented by their soft clustering assignments, i.e., cosine similarity (equals to the dot product since $l_2$ normalized) between node representations and prototypes:
\begin{equation}
  \label{node_u}
  \boldsymbol{z}_u = [f'(\boldsymbol{x}_u) \cdot \boldsymbol{\mu}_1, f'(\boldsymbol{x}_u) \cdot \boldsymbol{\mu}_2, ..., f'(\boldsymbol{x}_u) \cdot \boldsymbol{\mu}_K] \in \mathbb{R}^K,
\end{equation}
\begin{equation}
  \label{node_v}
  \boldsymbol{z}_v = [f'(\boldsymbol{x}_v) \cdot \boldsymbol{\mu}_1, f'(\boldsymbol{x}_v) \cdot \boldsymbol{\mu}_2, ..., f'(\boldsymbol{x}_v) \cdot \boldsymbol{\mu}_K] \in \mathbb{R}^K,
\end{equation}
and the feature vector of $(u, v)$ is represented by:
\begin{equation}
  \label{edge_feat}
  \boldsymbol{z}_{u,v} = [\boldsymbol{z}_u || \boldsymbol{z}_v] \in \mathbb{R}^{2K},
\end{equation}
where $||$ denotes the concatenating operation. Then, we employ a two-layer MLP $h(\cdot)$ to predict the new edge weight $w_{u,v}^{pre}$:
\begin{equation}
  \label{w_pre}
  w_{u,v}^{pre} = \varphi(h(\boldsymbol{z}_{u,v})),
\end{equation}
where $\varphi(\cdot)$ is the Sigmoid function.

Although $w_{u,v}^{pre}$ can be learned end-to-end under the contrastive clustering loss, the learning process lacks explicit supervision. To this end, we further propose the empirical edge weight as evidence to guide learning. We first define the clustering confidence for node $u$:
\begin{equation}
  \label{conf_u}
  \mathrm{conf}_u = \max_{1 \le k \le K} f'(\boldsymbol{x}_u) \cdot \boldsymbol{\mu}_k,
\end{equation}
and the similarity between nodes $u$ and $v$:
\begin{equation}
  \label{sim_u_v}
  \mathrm{sim}(u,v) = f'(\boldsymbol{x}_u) \cdot f'(\boldsymbol{x}_v).
\end{equation}
Then the empirical edge weight is computed based on a simple intuition: if nodes $u$ and $v$ have higher clustering confidence, higher similarity, and are assigned to the same cluster, then edge $(u,v)$ is more reliable and should be set a higher empirical weight, formulated as:
\begin{equation}
  \label{w_emp}
  w_{u,v}^{emp} = \varphi((2 \cdot \mathrm{ind}(u, v)-1) \cdot \mathrm{conf}_u \cdot \mathrm{conf}_v \cdot \mathrm{sim}(u, v)),
\end{equation}
where $\mathrm{ind}(\cdot)$ indicates whether $u$ and $v$ are assigned to the same class (1 if true, else 0). To constrain the empirical edge weight in [0,1], we perform min-max normalization on $\mathrm{conf}_u$, $\mathrm{conf}_v$ and $\mathrm{sim}(u, v)$ across all edges, and employ the Sigmoid function $\varphi(\cdot)$. Moreover, if $u$ and $v$ are assigned to different classes, we flip the value of $\mathrm{sim}(u, v)$ by setting $\mathrm{sim}(u, v) = 1 - \mathrm{sim}(u, v)$.

Next, we treat the empirical edge weights as prior, and employ mean squared error (MSE) as the loss function to guide the edge learning, encouraging the predicted edge weights $w_{u,v}^{pre}$ to be close to the empirical edge weights $w_{u,v}^{emp}$:
\begin{equation}
  \label{edge_loss}
  \mathcal{L}_{E} = \frac{1}{\vert \mathcal{E} \vert} \sum_{(u, v) \in \mathcal{E}}(w_{u,v}^{pre}-w_{u,v}^{emp})^2.
\end{equation}

Finally, we build the predicted adjacency matrix $\mathbf{A}^{pre}$ with $w^{pre}$ of all edges, and the original superpixel adjacency matrix is momentum updated to stabilize the training, following:
\begin{equation}
  \label{adj_update}
  \mathbf{A} = \gamma \mathbf{A}+(1-\gamma)\mathbf{A}^{pre},
\end{equation}
where $\gamma \in (0, 1)$ is the momentum coefficient.

\subsection{Training Procedure}
The total loss of the proposed method is defined as the combination of the contrastive clustering loss and the edge learning loss:
\begin{equation}
  \label{total_loss}
  \mathcal{L} = \mathcal{L}_{CC} + \beta \mathcal{L}_{E} = \mathcal{L}_{NA} + \alpha \mathcal{L}_{PC} + \beta \mathcal{L}_{E},
\end{equation}
where $\alpha>0$ and $\beta>0$ are both weight hyperparameters.

We illustrate the training procedure of the proposed method in algorithm \ref{alg}. In each epoch, we first compute the empirical edge weights and the predicted edge weights to update the adjacency matrix and the corresponding symmetrically normalized matrix. Second, we perform forward-propagation, computing node representations, prototypes and loss functions. Third, network parameters are optimized. Fourth, under evaluation mode, we apply spherical K-means on the output of $f'(\cdot)$ to get the current superpixel clustering assignments and then update the prototypes. Finally, we transform the clustering result from the superpixel level to the pixel level by assigning each superpixel's label to its internal pixels.

\begin{algorithm}[!t]
  \caption{Algorithm of the Proposed Method}
  \label{alg}
  \begin{algorithmic}[1]
    \REQUIRE Superpixels $\{\boldsymbol{x}\}$; Adjacency matrix $\mathbf{A}$; Online network $f(\cdot)$; Target network $f'(\cdot)$.\\
    \ENSURE Clustering results $\mathbf{R}$.\\
    \STATE Forward $f'(\boldsymbol{x})$ by Eq. \eqref{ssgco_eq}, perform K-means, and compute $\{\boldsymbol{\mu}\}$ by Eq. \eqref{prototype}.
    \FOR{$t=1$ to $T$}
    \STATE Compute $\{w^{emp}\}, \{w^{pre}\}$ by Eqs. \eqref{w_emp} and \eqref{w_pre}, and compute $\mathcal{L}_E$ by Eq. \eqref{edge_loss}.
    \STATE Update $\mathbf{A}$ by Eq. \eqref{adj_update}.
    \STATE Update $\widehat{\mathbf{A}}$ based on the current $\mathbf{A}$ by Eq. \eqref{renorm}.
    \STATE Randomly sample pixels to get $\{\boldsymbol{x}^+\}$.
    \STATE Forward $f(\boldsymbol{x}), f'(\boldsymbol{x}^+)$ by Eq. \eqref{ssgco_eq}.
    \STATE Compute $\mathcal{L}_{NA}$ by Eq. \eqref{na}.
    \STATE Compute $\{\boldsymbol{\mu}\}, \{\boldsymbol{\mu}^+\}$ by Eq. \eqref{prototype}.
    \STATE Compute $\mathcal{L}_{PC}$ by Eq. \eqref{pc}.
    \STATE Compute $\mathcal{L}$ by Eq. \eqref{total_loss}.
    \STATE Update $f(\cdot)$ with SGD optimizer.
    \STATE Update $f'(\cdot)$ by Eq. \eqref{target_mom_update}.
    \STATE Evaluation mode, forward $f'(\boldsymbol{x})$, perform K-means to get $\mathbf{R}$, and compute $\{\boldsymbol{\mu}\}$ by Eq. \eqref{prototype}.
    \ENDFOR
    \STATE Transform $\mathbf{R}$ from the superpixel level to the pixel level.
    \RETURN $\mathbf{R}$
  \end{algorithmic}
\end{algorithm}

\section{Experiments\label{experiments}}
\subsection{Datasets}
We evaluated our method on four public HSI datasets, including \textit{Indian Pines}, \textit{Pavia University}, \textit{Botswana} and \textit{Trento}.

\textit{1) Indian Pines:} This dataset uses the AVIRIS sensor and has a resolution of 145 $\times$ 145 with 200 spectral bands after removing 24 water absorption bands. It includes 16 different classes with a total of 10249 labeled pixels.

\textit{2) Pavia University:} Utilizing the ROSIS sensor, this dataset has a resolution of 610 $\times$ 340 and 103 spectral bands. It comprises 9 classes and has 42776 labeled pixels.

\textit{3) Botswana:} This dataset is captured with the Hyperion sensor, featuring a high resolution of 1476 $\times$ 256 and 145 spectral bands. It contains 14 classes and a total of 3248 labeled pixels.

\textit{4) Trento:} With the AISA Eagle sensor, this dataset offers a resolution of 600 $\times$ 166 and includes 63 spectral bands. It has 6 classes and 30214 labeled pixels.

\begin{table*}[t]
\centering
\caption{Clustering Performance on HSI Datasets, \textit{Indian Pines}, \textit{Pavia University}, \textit{Botswana} and \textit{Trento} are Abbreviated to \textit{IP}, \textit{PU}, \textit{BO} and \textit{TR}, $\bullet$ Indicates that our Method is Statistically Superior to the Compared Method According to the Pairwise $t$-Test at 0.05 Significance Level.\label{clustering_res}}
\resizebox{\textwidth}{!}{
\renewcommand\arraystretch{1.14}
\begin{tabular}{c!{\vrule width \lightrulewidth}c!{\vrule width \lightrulewidth}cccccccccccc} 
\toprule
Dataset                      & Metric (\%)   & \begin{tabular}[c]{@{}c@{}}K-means\cite{kmeans} \\ (TPAMI 02)\end{tabular} & \begin{tabular}[c]{@{}c@{}}SSC\cite{ssc} \\ (TPAMI 13)\end{tabular} & \begin{tabular}[c]{@{}c@{}}NCSC\cite{ncsc} \\ (TGRS 22)\end{tabular} & \begin{tabular}[c]{@{}c@{}}SGLSC\cite{sglsc} \\ (TGRS 22)\end{tabular} & \begin{tabular}[c]{@{}c@{}}DGAE\cite{dgae} \\ (TCSVT 22)\end{tabular} & \begin{tabular}[c]{@{}c@{}}EGRC\cite{EGRC} \\(TIP 23)\end{tabular} & \begin{tabular}[c]{@{}c@{}}S$^3$-ULDA\cite{s3ulda} \\ (TGRS 23)\end{tabular} & \begin{tabular}[c]{@{}c@{}}S$^2$DL\cite{s2dl} \\ (TGRS 24)\end{tabular} & \begin{tabular}[c]{@{}c@{}}SDST\cite{sdst} \\ (TGRS 24)\end{tabular} & \begin{tabular}[c]{@{}c@{}}SPGCC\cite{spgcc} \\ (TCSVT 24)\end{tabular} & \begin{tabular}[c]{@{}c@{}}SAPC\cite{sapc} \\ (TCSVT 24)\end{tabular} & Ours                 \\ 
\midrule
\multirow{8}{*}{\textit{IP}} & ACC           & 34.72±1.00$_\bullet$                                          & 47.99±0.46$_\bullet$                                      & 53.48±0.84$_\bullet$                                      & 58.29±1.19$_\bullet$                                       & 57.10±3.39$_\bullet$                                       & 56.02±1.87$_\bullet$                                       & 65.41±3.37$_\bullet$                                            & \uline{64.70±0.00}$_\bullet$                                 & 50.58±4.46$_\bullet$                                      & 67.40±1.17$_\bullet$                                        & 55.88±0.00$_\bullet$                                       & \textbf{70.01±1.95}  \\
                             & $\mathcal{K}$ & 28.39±1.10$_\bullet$                                          & 42.13±0.57$_\bullet$                                      & 48.54±1.12$_\bullet$                                      & 54.54±1.22$_\bullet$                                       & 52.18±3.56$_\bullet$                                       & 49.80±2.51$_\bullet$                                       & 61.10±3.50$_\bullet$                                            & 60.24±0.00$_\bullet$                                         & 44.01±4.27$_\bullet$                                      & \uline{63.62±1.34}$_\bullet$                                & 49.59±0.00$_\bullet$                                       & \textbf{66.55±2.12}  \\
                             & NMI           & 43.24±0.57$_\bullet$                                          & 57.81±0.44$_\bullet$                                      & 59.47±1.40$_\bullet$                                      & 63.65±0.35$_\bullet$                                       & 63.68±1.03$_\bullet$                                       & 56.91±1.69$_\bullet$                                       & 67.65±1.65$_\bullet$                                            & \textbf{71.97±0.00}$\phantom{_\bullet}$                      & 50.21±1.80$_\bullet$                                      & 67.54±1.26$_\bullet$                                        & 54.03±0.00$_\bullet$                                       & \uline{71.17±1.31}   \\
                             & ARI           & 21.02±0.99$_\bullet$                                          & 31.46±0.55$_\bullet$                                      & 36.13±1.42$_\bullet$                                      & 40.21±0.49$_\bullet$                                       & 45.35±4.75$_\bullet$                                       & 38.72±3.50$_\bullet$                                       & 48.99±3.76$_\bullet$                                            & 52.93±0.00$_\bullet$                                         & 31.51±3.83$_\bullet$                                      & \textbf{56.02±1.94}$\phantom{_\bullet}$                     & 36.51±0.00$_\bullet$                                       & \uline{55.99±2.07}   \\
                             & Precision     & 37.74±1.15$_\bullet$                                          & 43.93±1.72$_\bullet$                                      & 42.74±2.88$_\bullet$                                      & 51.88±1.62$_\bullet$                                       & 50.59±1.81$_\bullet$                                       & 42.50±4.87$_\bullet$                                       & 51.34±1.87$_\bullet$                                            & \textbf{59.63±0.00}$\phantom{_\bullet}$                      & 45.34±2.71$_\bullet$                                      & 52.66±2.19$_\bullet$                                        & 49.61±0.00$_\bullet$                                       & \uline{58.72±2.23}   \\
                             & Recall        & 40.27±0.38$_\bullet$                                          & 47.42±5.06$_\bullet$                                      & 51.12±0.62$_\bullet$                                      & 57.11±1.97$_\bullet$                                       & 49.79±3.89$_\bullet$                                       & 46.36±4.65$_\bullet$                                       & 51.11±3.16$_\bullet$                                            & 59.08±0.00$_\bullet$                                         & 46.37±1.38$_\bullet$                                      & \uline{61.34±3.48}$_\bullet$                                & 52.39±0.00$_\bullet$                                       & \textbf{65.77±4.42}  \\
                             & F1            & 34.24±0.87$_\bullet$                                          & 40.65±1.34$_\bullet$                                      & 40.76±0.92$_\bullet$                                      & 49.50±1.76$_\bullet$                                       & 46.00±2.58$_\bullet$                                       & 39.26±3.12$_\bullet$                                       & 48.07±3.10$_\bullet$                                            & \uline{55.19±0.00}$\phantom{_\bullet}$                       & 44.30±2.74$_\bullet$                                      & 52.30±1.89$_\bullet$                                        & 44.67±0.00$_\bullet$                                       & \textbf{57.65±2.38}  \\
                             & Purity        & 52.26±0.87$_\bullet$                                          & 57.78±0.51$_\bullet$                                      & 62.87±0.65$_\bullet$                                      & 69.58±1.25$_\bullet$                                       & 68.94±1.51$_\bullet$                                       & 59.78±2.92$_\bullet$                                       & 72.52±2.06$_\bullet$                                            & \uline{75.32±0.00}$_\bullet$                                 & 56.51±1.36$_\bullet$                                      & 73.94±1.78$_\bullet$                                        & 59.33±0.00$_\bullet$                                       & \textbf{77.18±1.03}  \\ 
\midrule
\multirow{8}{*}{\textit{PU}} & ACC           & 51.15±0.02$_\bullet$                                          & 55.17±0.00$_\bullet$                                      & 56.08±3.69$_\bullet$                                      & 57.24±0.34$_\bullet$                                       & 58.56±5.21$\phantom{_\bullet}$                             & 59.91±3.00$_\bullet$                                       & 59.42±0.28$_\bullet$                                            & 60.13±0.00$_\bullet$                                         & 61.03±1.80$_\bullet$                                      & 62.62±0.38$_\bullet$                                        & \uline{63.94±0.00}$_\bullet$                               & \textbf{70.00±3.46}  \\
                             & $\mathcal{K}$ & 41.49±0.06$_\bullet$                                          & 45.57±0.00$_\bullet$                                      & 44.89±5.06$_\bullet$                                      & 48.28±0.36$_\bullet$                                       & 49.15±5.07$_\bullet$                                       & 47.08±2.91$_\bullet$                                       & 50.98±0.47$_\bullet$                                            & 49.06±0.00$_\bullet$                                         & 50.18±2.12$_\bullet$                                      & \uline{54.82±0.46}$_\bullet$                                & 53.02±0.00$_\bullet$                                       & \textbf{62.29±3.65}  \\
                             & NMI           & 53.97±0.01$_\bullet$                                          & 51.48±0.00$_\bullet$                                      & 47.34±4.27$_\bullet$                                      & 53.69±0.39$_\bullet$                                       & 53.72±2.58$_\bullet$                                       & 54.59±3.71$_\bullet$                                       & \textbf{62.93±0.08}$\phantom{_\bullet}$                         & 48.15±0.00$_\bullet$                                         & 56.41±1.44$_\bullet$                                      & 57.84±1.20$\phantom{_\bullet}$                              & 61.89±0.00$\phantom{_\bullet}$                             & \uline{61.90±3.18}   \\
                             & ARI           & 31.02±0.03$_\bullet$                                          & 35.26±0.00$_\bullet$                                      & 42.84±5.04$_\bullet$                                      & 42.64±0.20$_\bullet$                                       & 43.65±6.69$\phantom{_\bullet}$                             & 43.10±2.61$_\bullet$                                       & \uline{49.04±0.18}$\phantom{_\bullet}$                          & 43.12±0.00$_\bullet$                                         & 42.94±2.63$_\bullet$                                      & 46.19±0.89$\phantom{_\bullet}$                              & 44.91±0.00$_\bullet$                                       & \textbf{57.26±8.57}  \\
                             & Precision     & 51.83±0.01$_\bullet$                                          & 43.00±0.00$_\bullet$                                      & 40.91±4.93$_\bullet$                                      & 50.29±0.39$_\bullet$                                       & 48.16±4.44$_\bullet$                                       & 46.73±5.54$_\bullet$                                       & \uline{56.77±0.70}$\phantom{_\bullet}$                          & 48.68±0.00$_\bullet$                                         & 53.97±3.11$\phantom{_\bullet}$                            & 56.32±1.16$\phantom{_\bullet}$                              & 56.21±0.00$\phantom{_\bullet}$                             & \textbf{57.82±1.22}  \\
                             & Recall        & 53.59±0.01$_\bullet$                                          & 48.81±0.00$_\bullet$                                      & 46.19±7.95$_\bullet$                                      & 53.75±0.35$_\bullet$                                       & 50.75±3.47$_\bullet$                                       & 49.13±6.92$_\bullet$                                       & 55.63±0.51$_\bullet$                                            & 50.37±0.00$_\bullet$                                         & 55.34±1.58$_\bullet$                                      & 58.96±0.42$\phantom{_\bullet}$                              & \uline{59.20±0.00}$\phantom{_\bullet}$                     & \textbf{60.55±3.09}  \\
                             & F1            & 50.34±0.01$_\bullet$                                          & 41.29±0.00$_\bullet$                                      & 40.09±5.01$_\bullet$                                      & 48.17±0.38$_\bullet$                                       & 45.17±2.27$_\bullet$                                       & 43.76±5.71$_\bullet$                                       & 53.21±0.53$_\bullet$                                            & 40.21±0.00$_\bullet$                                         & 51.08±2.06$_\bullet$                                      & \uline{56.05±0.99}$\phantom{_\bullet}$                      & 56.02±0.00$\phantom{_\bullet}$                             & \textbf{57.18±2.09}  \\
                             & Purity        & 69.86±0.00$_\bullet$                                          & 63.87±0.00$_\bullet$                                      & 64.63±2.20$_\bullet$                                      & 71.88±0.20$_\bullet$                                       & 71.17±1.99$_\bullet$                                       & 68.39±2.92$_\bullet$                                       & 76.37±0.47$\phantom{_\bullet}$                                  & 68.25±0.00$_\bullet$                                         & 69.80±1.66$_\bullet$                                      & \uline{77.06±0.62}$\phantom{_\bullet}$                      & 71.32±0.00$_\bullet$                                       & \textbf{77.19±2.79}  \\ 
\midrule
\multirow{8}{*}{\textit{BO}} & ACC           & 60.58±0.13$_\bullet$                                          & 58.54±1.01$_\bullet$                                      & 47.23±4.27$_\bullet$                                      & 58.68±0.01$_\bullet$                                       & 58.48±1.12$_\bullet$                                       & 62.83±5.75$_\bullet$                                       & 45.95±3.74$_\bullet$                                            & 60.53±0.00$_\bullet$                                         & \uline{72.70±3.18}$\phantom{_\bullet}$                    & 62.18±3.45$_\bullet$                                        & 66.10±0.00$_\bullet$                                       & \textbf{77.66±1.28}  \\
                             & $\mathcal{K}$ & 57.40±0.15$_\bullet$                                          & 54.93±1.12$_\bullet$                                      & 42.68±4.63$_\bullet$                                      & 55.06±0.01$_\bullet$                                       & 55.09±1.22$_\bullet$                                       & 59.63±6.28$_\bullet$                                       & 41.46±4.06$_\bullet$                                            & 57.19±0.00$_\bullet$                                         & \uline{70.40±3.48}$\phantom{_\bullet}$                    & 59.14±3.73$_\bullet$                                        & 63.17±0.00$_\bullet$                                       & \textbf{75.79±1.39}  \\
                             & NMI           & 69.19±0.01$_\bullet$                                          & 74.23±0.10$_\bullet$                                      & 64.14±1.88$_\bullet$                                      & 67.12±0.18$_\bullet$                                       & 68.47±1.34$_\bullet$                                       & 73.76±3.93$_\bullet$                                       & 62.40±2.86$_\bullet$                                            & 68.01±0.00$_\bullet$                                         & \uline{76.02±2.15}$_\bullet$                              & 73.48±1.10$_\bullet$                                        & 71.57±0.00$_\bullet$                                       & \textbf{82.90±1.30}  \\
                             & ARI           & 50.08±0.03$_\bullet$                                          & 52.91±0.63$_\bullet$                                      & 38.51±4.18$_\bullet$                                      & 48.08±0.39$_\bullet$                                       & 44.85±1.26$_\bullet$                                       & 54.34±4.30$_\bullet$                                       & 35.46±3.93$_\bullet$                                            & 43.76±0.00$_\bullet$                                         & \uline{58.70±3.66}$_\bullet$                              & 54.06±2.87$_\bullet$                                        & 48.15±0.00$_\bullet$                                       & \textbf{69.03±2.38}  \\
                             & Precision     & 58.51±0.07$_\bullet$                                          & 53.43±1.93$_\bullet$                                      & 43.22±7.31$_\bullet$                                      & 60.02±3.43$_\bullet$                                       & 57.02±1.38$_\bullet$                                       & 53.14±9.53$_\bullet$                                       & 41.68±4.58$_\bullet$                                            & 67.13±0.00$_\bullet$                                         & \uline{72.81±4.06}$\phantom{_\bullet}$                    & 61.06±4.54$_\bullet$                                        & 67.66±0.00$_\bullet$                                       & \textbf{73.90±2.42}  \\
                             & Recall        & 62.54±0.13$_\bullet$                                          & 55.47±1.56$_\bullet$                                      & 46.23±3.84$_\bullet$                                      & 55.30±0.03$_\bullet$                                       & 60.88±1.24$_\bullet$                                       & 60.31±6.37$_\bullet$                                       & 47.71±3.40$_\bullet$                                            & 60.50±0.00$_\bullet$                                         & \uline{70.82±3.34}$\phantom{_\bullet}$                    & 61.44±3.81$_\bullet$                                        & 66.81±0.00$_\bullet$                                       & \textbf{77.65±2.39}  \\
                             & F1            & 58.19±0.09$_\bullet$                                          & 51.05±1.77$_\bullet$                                      & 41.29±4.95$_\bullet$                                      & 53.03±0.14$_\bullet$                                       & 56.35±1.36$_\bullet$                                       & 53.17±8.22$_\bullet$                                       & 39.24±3.37$_\bullet$                                            & 60.06±0.00$_\bullet$                                         & \uline{69.40±3.12}$\phantom{_\bullet}$                    & 59.04±4.17$_\bullet$                                        & 64.41±0.00$_\bullet$                                       & \textbf{74.27±2.52}  \\
                             & Purity        & 64.84±0.12$_\bullet$                                          & 62.03±0.96$_\bullet$                                      & 50.05±4.82$_\bullet$                                      & 59.57±0.01$_\bullet$                                       & 63.77±1.12$_\bullet$                                       & 64.05±6.06$_\bullet$                                       & 48.01±2.83$_\bullet$                                            & 60.53±0.00$_\bullet$                                         & \uline{73.31±2.33}$_\bullet$                              & 66.40±2.20$_\bullet$                                        & 66.10±0.00$_\bullet$                                       & \textbf{77.71±1.32}  \\ 
\midrule
\multirow{8}{*}{\textit{TR}} & ACC           & 64.83±0.02$_\bullet$                                          & 79.18±0.05$_\bullet$                                      & 75.70±2.13$_\bullet$                                      & 82.75±0.27$_\bullet$                                       & 86.50±0.00$_\bullet$                                       & 77.73±1.93$_\bullet$                                       & 75.47±0.00$_\bullet$                                            & 86.14±0.00$_\bullet$                                         & 87.97±0.74$_\bullet$                                      & 89.67±3.53$\phantom{_\bullet}$                              & \uline{89.96±0.00}$_\bullet$                               & \textbf{93.11±0.96}  \\
                             & $\mathcal{K}$ & 55.63±0.03$_\bullet$                                          & 72.57±0.07$_\bullet$                                      & 69.00±2.60$_\bullet$                                      & 77.33±0.35$_\bullet$                                       & 82.13±0.00$_\bullet$                                       & 69.32±2.75$_\bullet$                                       & 68.62±0.06$_\bullet$                                            & 81.27±0.00$_\bullet$                                         & 83.94±0.91$_\bullet$                                      & 86.38±4.61$\phantom{_\bullet}$                              & \uline{86.62±0.00}$_\bullet$                               & \textbf{90.82±1.26}  \\
                             & NMI           & 51.87±0.01$_\bullet$                                          & 66.73±0.03$_\bullet$                                      & 72.78±2.38$_\bullet$                                      & 78.42±0.06$_\bullet$                                       & 75.62±0.00$_\bullet$                                       & 75.54±0.81$_\bullet$                                       & 77.90±0.01$_\bullet$                                            & 79.80±0.00$_\bullet$                                         & 76.15±1.49$_\bullet$                                      & 85.46±1.17$_\bullet$                                        & \uline{86.74±0.00}$_\bullet$                               & \textbf{89.74±0.94}  \\
                             & ARI           & 36.82±0.02$_\bullet$                                          & 55.56±0.09$_\bullet$                                      & 71.92±2.15$_\bullet$                                      & 78.27±0.20$_\bullet$                                       & 79.47±0.00$_\bullet$                                       & 73.90±1.91$_\bullet$                                       & 72.81±0.00$_\bullet$                                            & 86.44±0.00$_\bullet$                                         & 82.42±1.91$_\bullet$                                      & 89.34±2.26$_\bullet$                                        & \uline{91.69±0.00}$_\bullet$                               & \textbf{94.00±1.30}  \\
                             & Precision     & 59.63±0.01$_\bullet$                                          & 80.26±0.02$_\bullet$                                      & 64.54±4.21$_\bullet$                                      & 63.69±0.42$_\bullet$                                       & 68.48±0.00$_\bullet$                                       & 51.10±2.14$_\bullet$                                       & 55.46±0.00$_\bullet$                                            & 68.19±0.00$_\bullet$                                         & 69.75±0.72$_\bullet$                                      & 73.89±7.29$_\bullet$                                        & \uline{84.31±0.00}$_\bullet$                               & \textbf{93.25±0.67}  \\
                             & Recall        & 58.24±0.01$_\bullet$                                          & 83.02±0.02$_\bullet$                                      & 60.56±2.65$_\bullet$                                      & 65.29±0.32$_\bullet$                                       & 69.84±0.00$_\bullet$                                       & 55.75±3.76$_\bullet$                                       & 59.22±0.00$_\bullet$                                            & 63.60±0.00$_\bullet$                                         & 70.15±0.04$_\bullet$                                      & 76.64±5.06$_\bullet$                                        & \uline{83.09±0.00}$_\bullet$                               & \textbf{93.11±0.96}  \\
                             & F1            & 55.25±0.02$_\bullet$                                          & 78.78±0.03$_\bullet$                                      & 61.57±3.22$_\bullet$                                      & 60.65±0.49$_\bullet$                                       & 68.75±0.00$_\bullet$                                       & 49.50±4.57$_\bullet$                                       & 55.81±0.00$_\bullet$                                            & 62.83±0.00$_\bullet$                                         & 69.73±0.44$_\bullet$                                      & 74.31±6.71$_\bullet$                                        & \uline{82.88±0.00}$_\bullet$                               & \textbf{92.61±0.95}  \\
                             & Purity        & 65.61±0.01$_\bullet$                                          & 79.18±0.05$_\bullet$                                      & 87.27±1.68$_\bullet$                                      & 82.94±0.27$_\bullet$                                       & 87.09±0.00$_\bullet$                                       & 77.86±2.00$_\bullet$                                       & 86.87±0.00$_\bullet$                                            & 88.08±0.00$_\bullet$                                         & 88.66±0.81$_\bullet$                                      & \uline{91.75±2.81}$\phantom{_\bullet}$                      & 89.96±0.00$_\bullet$                                       & \textbf{93.66±0.74}  \\
\bottomrule
\end{tabular}}
\end{table*}

\begin{figure*}[t]
  \centering
  \hfill
  \subfloat[]{\includegraphics[width=0.076\textwidth]{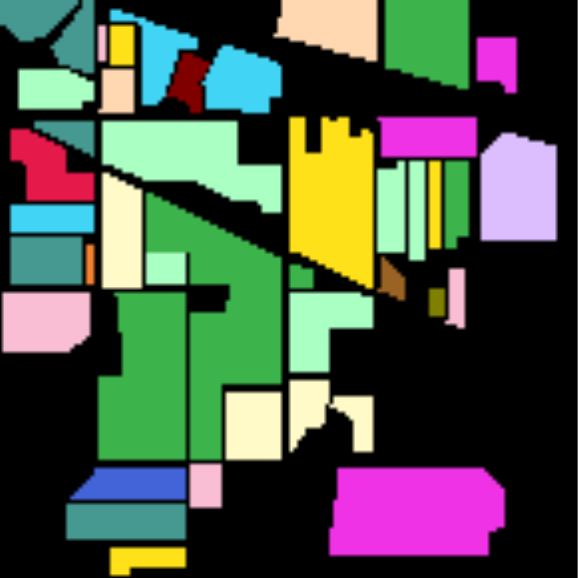}\label{IP_gt}}
  \hfill
  \subfloat[]{\includegraphics[width=0.076\textwidth]{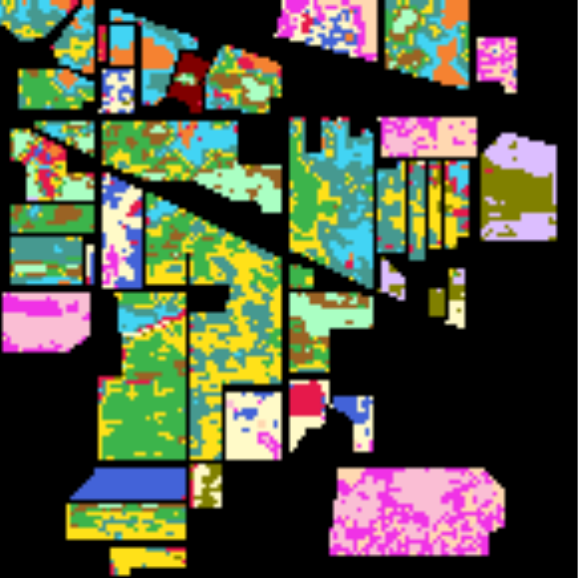}\label{IP_kmeans}}
  \hfill
  \subfloat[]{\includegraphics[width=0.076\textwidth]{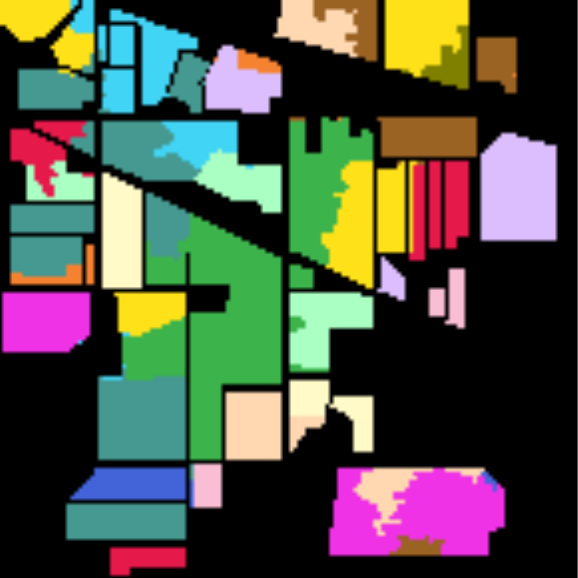}\label{IP_ssc}}
  \hfill
  \subfloat[]{\includegraphics[width=0.076\textwidth]{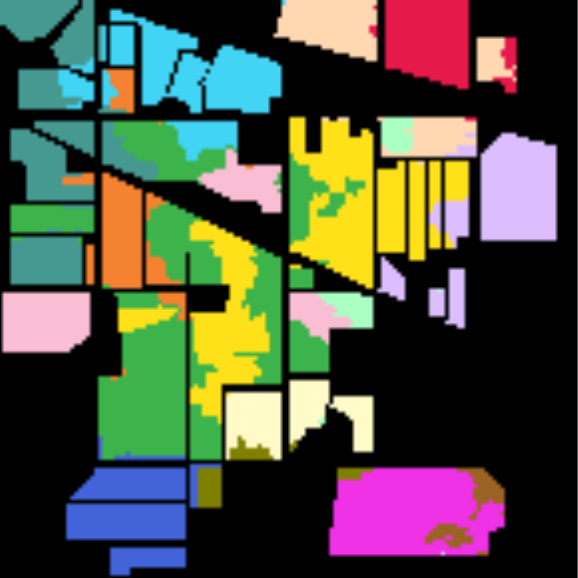}\label{IP_ncsc}}
  \hfill
  \subfloat[]{\includegraphics[width=0.076\textwidth]{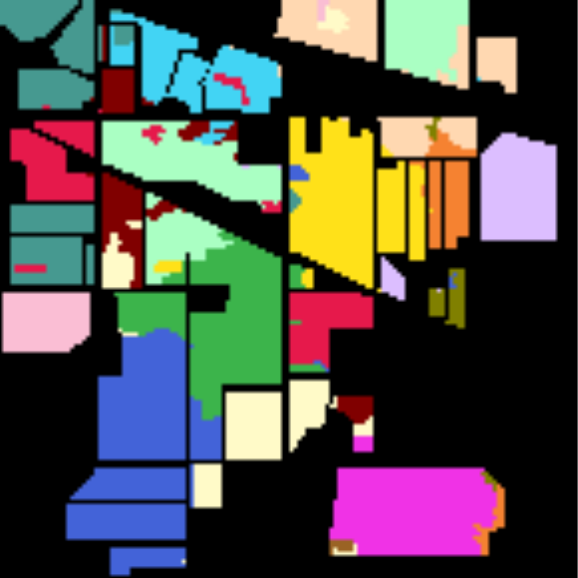}\label{IP_sglsc}}
  \hfill
  \subfloat[]{\includegraphics[width=0.076\textwidth]{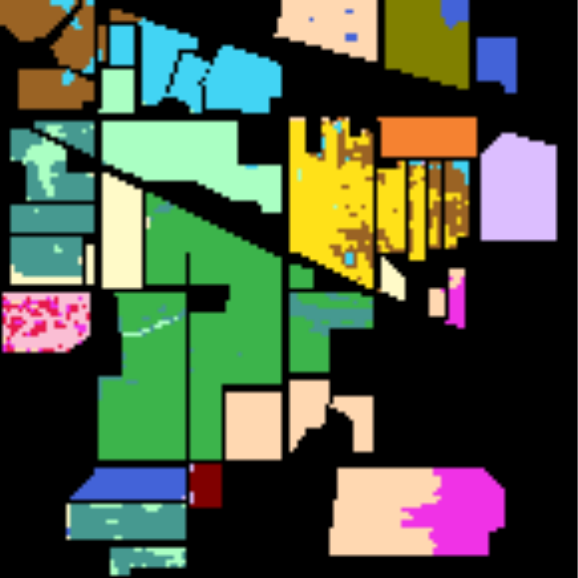}\label{IP_dgae}}
  \hfill
  \subfloat[]{\includegraphics[width=0.076\textwidth]{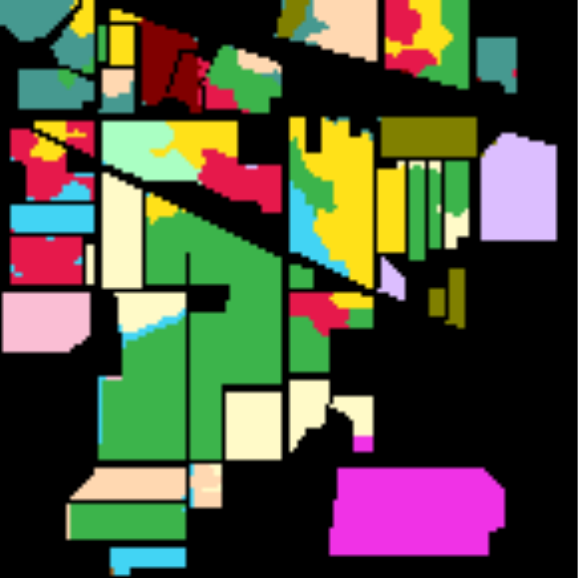}\label{IP_egrc}}
  \hfill
  \subfloat[]{\includegraphics[width=0.076\textwidth]{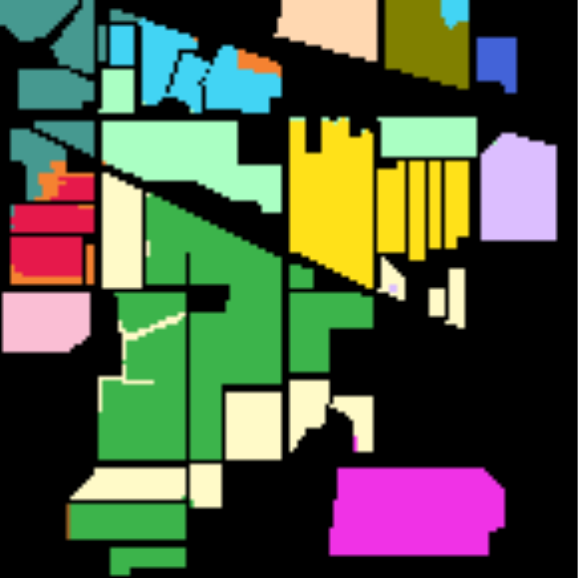}\label{IP_s3ulda}}
  \hfill
  \subfloat[]{\includegraphics[width=0.076\textwidth]{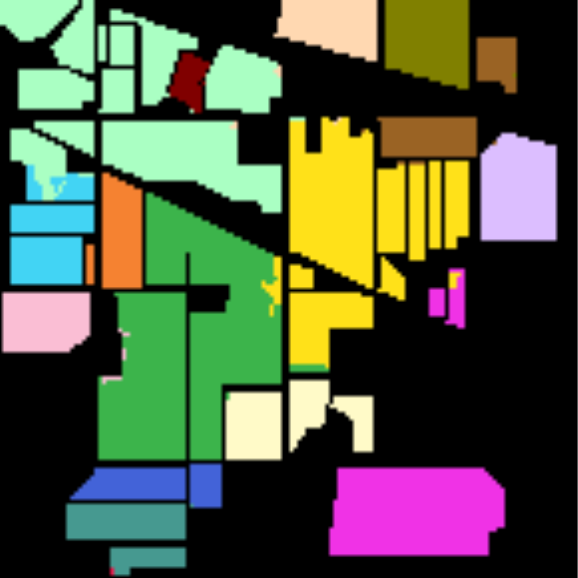}\label{IP_s2dl}}
  \hfill
  \subfloat[]{\includegraphics[width=0.076\textwidth]{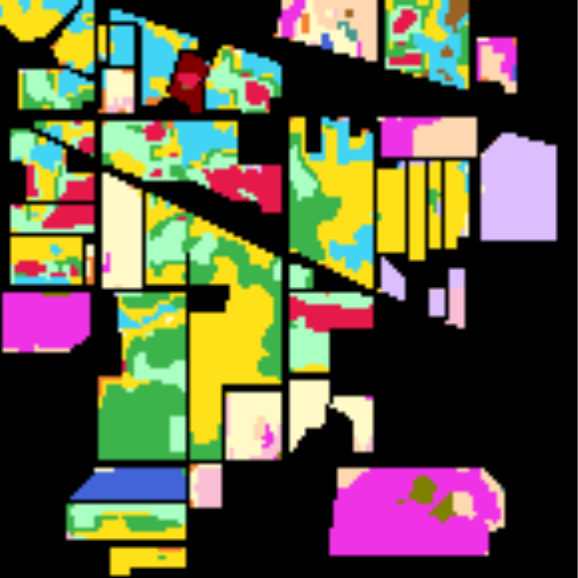}\label{IP_sdst}}
  \hfill
  \subfloat[]{\includegraphics[width=0.076\textwidth]{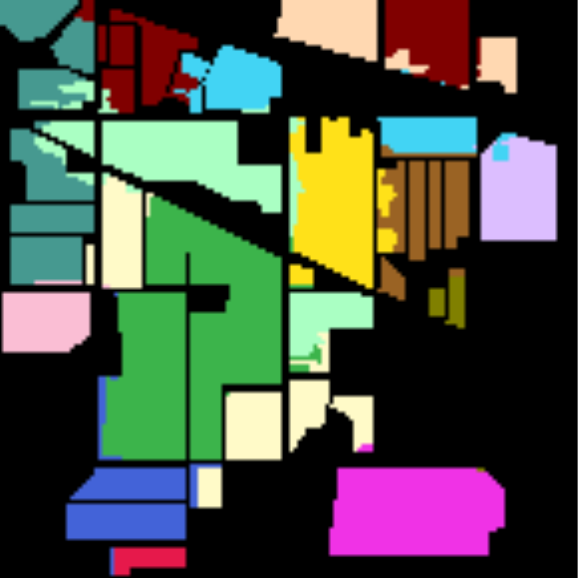}\label{IP_spgcc}}
  \hfill
  \subfloat[]{\includegraphics[width=0.076\textwidth]{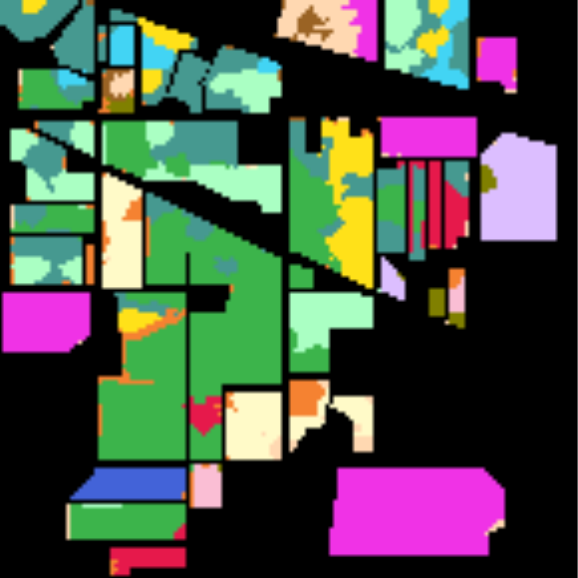}\label{IP_sapc}}
  \hfill
  \subfloat[]{\includegraphics[width=0.076\textwidth]{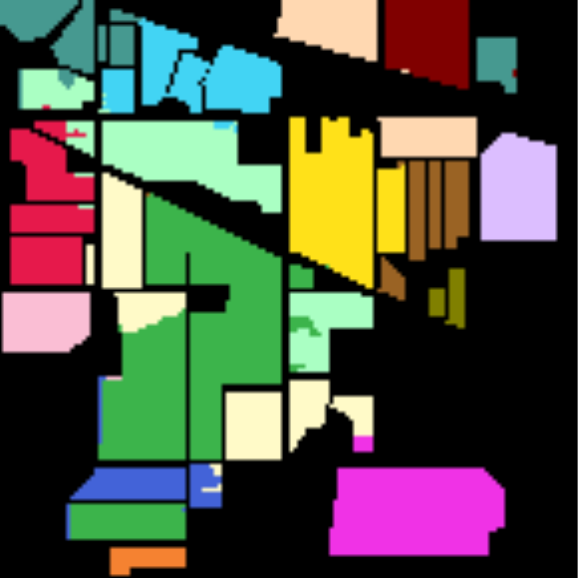}\label{IP_our}}
  \hfill
  \caption{Clustering maps on \textit{Indian Pines}: (a) ground truth, (b) K-means, (c) SSC, (d) NCSC, (e) SGLSC, (f) DGAE, (g) EGRC, (h) S$^3$-ULDA, (i) S$^2$DL, (j) SDST, (k) SPGCC, (l) SAPC, (m) Ours.}
  \label{IP_clustering_map}
\end{figure*}

\subsection{Experiment Setup}
\textit{1) Implementation Details:} The input HSI is standardized across each spectral band. To decrease computational costs, PCA is adopted to reduce the number of bands to 40, 20, 25 and 40 on \textit{Indian Pines}, \textit{Pavia University}, \textit{Botswana} and \textit{Trento}, respectively. In the contrastive clustering framework, the hidden size of the predictor $g(\cdot)$ is set to 512. The momentum coefficient $m$ for updating $f'(\cdot)$ is set to 0.99. For neighborhood alignment, $\sigma$ is set to $1 \times 10^{-3}$. For prototype contrast, $\tau$ is set to 0.7. In the structural-spectral graph convolutional operator, the kernel size of 1D convolution in the first layer is set to 7, with subsequent layers decreasing by 2, but not smaller than 3. The output channels in the first layer is set to 8, with subsequent layers multiplying by 2, but not larger than 64. FC layers in graph convolution keep output dimensions equal to input dimensions. In the evidence-guided adaptive edge learning, $h(\cdot)$ consists of a two-layer MLP with the hidden size of $K$. We used the SGD optimizer with the learning rate, weight decay, and momentum coefficient set to 0.05, $5 \times 10^{-4}$, and 0.9, respectively. In addition, we adopted cosine annealing and set the learning rate of $g(\cdot)$ $10 \times$ as that of other modules following \cite{byol}. We repeated all experiments five times and reported the mean and standard deviation of the results. We carried out all experiments in a Linux server with an AMD EPYC 7352 CPU and an NVIDIA RTX 4090 GPU.

\textit{2) Compared Methods:} To thoroughly assess the clustering performance of the proposed method, we compare it against eleven baseline methods, including K-means \cite{kmeans}, SSC \cite{ssc}, NCSC \cite{ncsc}, SGLSC \cite{sglsc}, DGAE \cite{dgae}, EGRC \cite{EGRC}, S$^3$-ULDA \cite{s3ulda}, S$^2$DL \cite{s2dl}, SDST \cite{sdst}, SPGCC \cite{spgcc} and SAPC \cite{sapc}. For K-means, clustering is conducted at the pixel level. In contrast, SSC performs clustering at the superpixel level to mitigate the high memory cost. For other methods, we follow the settings in their papers and exhaustively search optimal values for hyperparameters. Here we briefly introduce each method:

\textit{K-means} \cite{kmeans} is a traditional centroid-based clustering algorithm with ``K-means++" initialization.

\textit{SSC} \cite{ssc} is a traditional subspace-based clustering algorithm with the sparse constraint and considering the outliers.

\textit{NCSC} \cite{ncsc} introduces neighborhood contrastive regularization into deep subspace clustering.

\textit{SGLSC} \cite{sglsc} constructs global and local similarity graphs and performs subspace clustering.

\textit{DGAE} \cite{dgae} builds superpixel spatial and spectral graph and proposes a dual graph autoencoder.

\textit{EGRC} \cite{EGRC} refines the initial graph by adding edges between the nearest node pairs based on nodes' current embeddings and is trained under the autoencoder framework.

\textit{S$^3$-ULDA} \cite{s3ulda} is a superpixel dimensionality reduction method based on LDA, followed by K-means.

\textit{S$^2$DL} \cite{s2dl} performs diffusion learning based on superpixels and spatial regularization to achieve geometry-based clustering.

\textit{SDST} \cite{sdst} proposes a double-structure transformer with shared autoformer and achieves clustering under the autoencoder framework.

\textit{SPGCC} \cite{spgcc} proposes superpixel graph contrastive clustering with semantic-invariant augmentations.

\textit{SAPC} \cite{sapc} constructs pixel-anchor and anchor-anchor graphs and proposes structured anchor projected clustering.

\textit{3) Evaluation Metrics:} To quantitatively evaluate the clustering performance of different methods, we adopt eight mainstream clustering metrics: accuracy (ACC), kappa coefficient ($\mathcal{K}$), normalized mutual information (NMI), adjusted Rand index (ARI), Precision, Recall, F1-score (F1) and Purity. For Precision, Recall and F1, we use the macro-average form. Among these metrics, ACC, NMI, Precision, Recall, F1 and Purity range in [0, 1] while Kappa and ARI range in [-1, 1]. Each metric is reported as a percentage, with a higher value indicating the better clustering result. Following the common process, we adopt the Hungarian algorithm \cite{hungarian} to map the predicted labels based on the ground truth before calculating clustering metrics.

\begin{figure*}[t]
  \centering
  \hfill
  \subfloat[]{\includegraphics[width=0.247\textwidth]{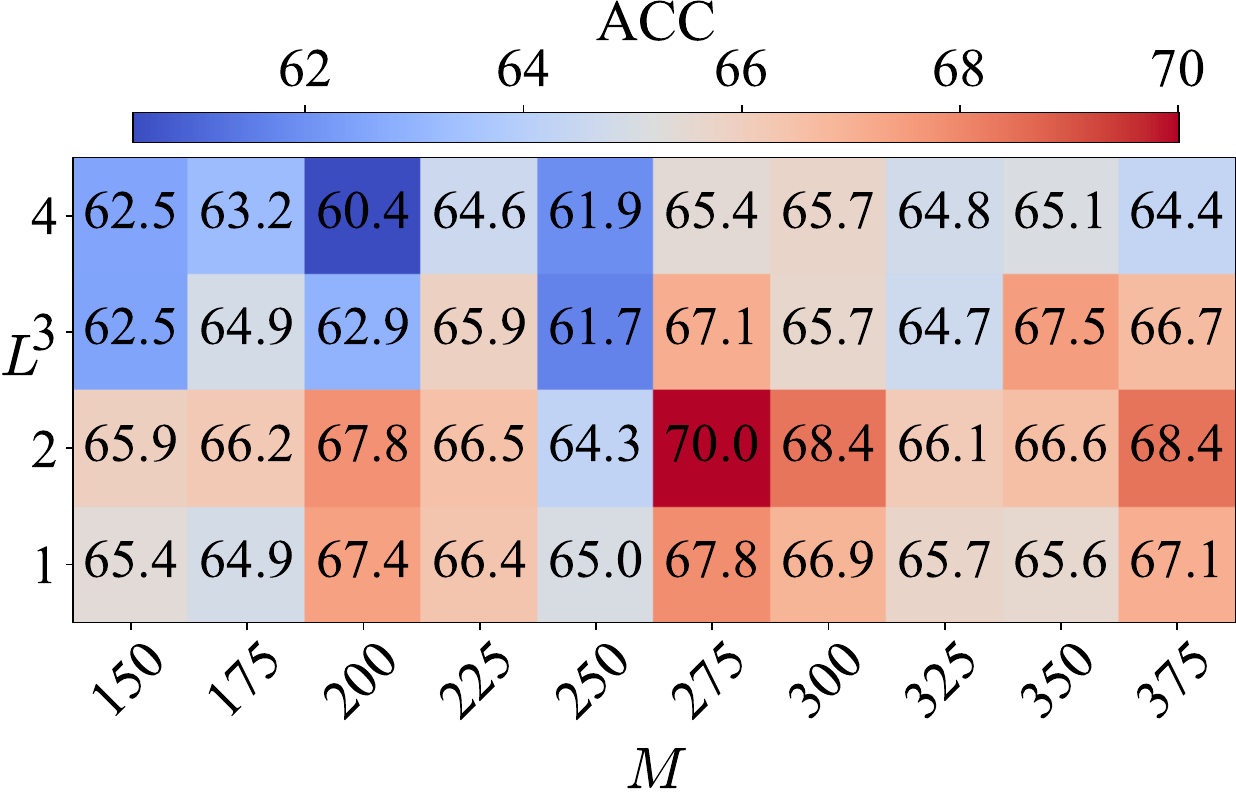}}
  \hfill
  \subfloat[]{\includegraphics[width=0.247\textwidth]{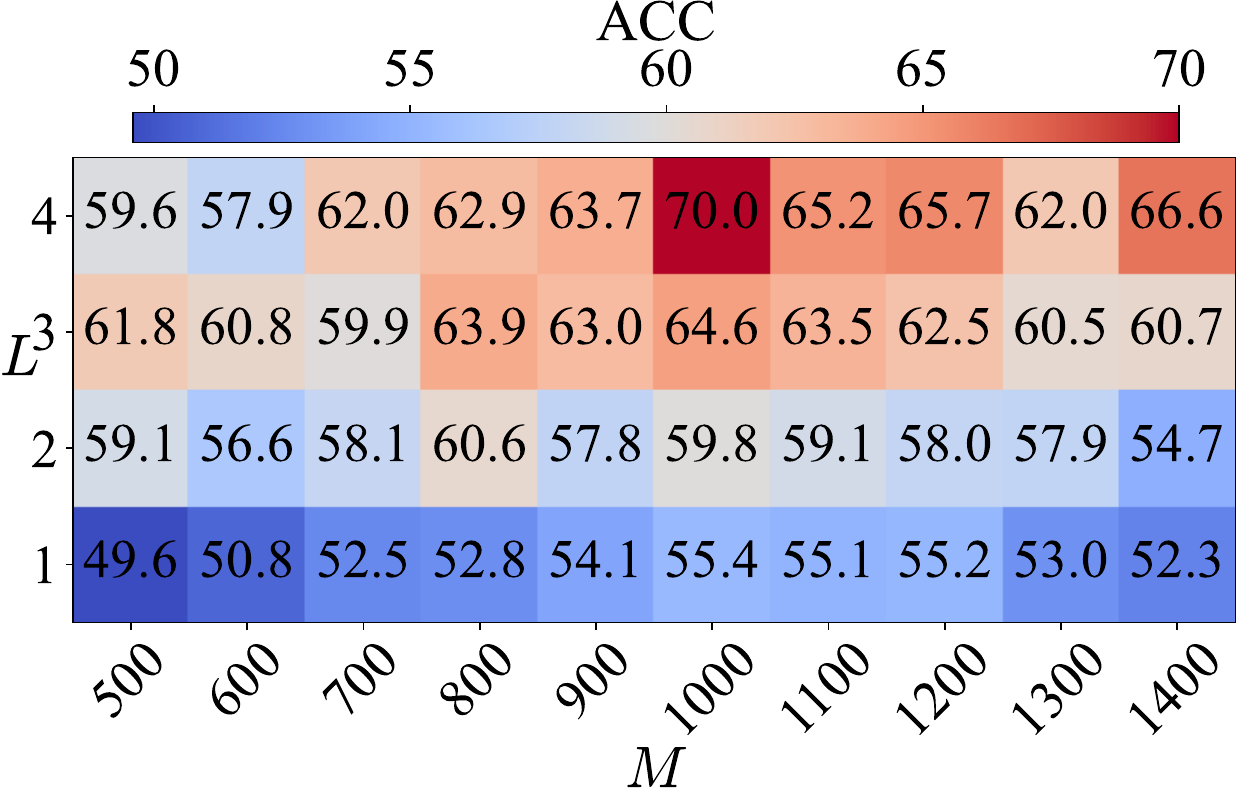}}
  \hfill
  \subfloat[]{\includegraphics[width=0.247\textwidth]{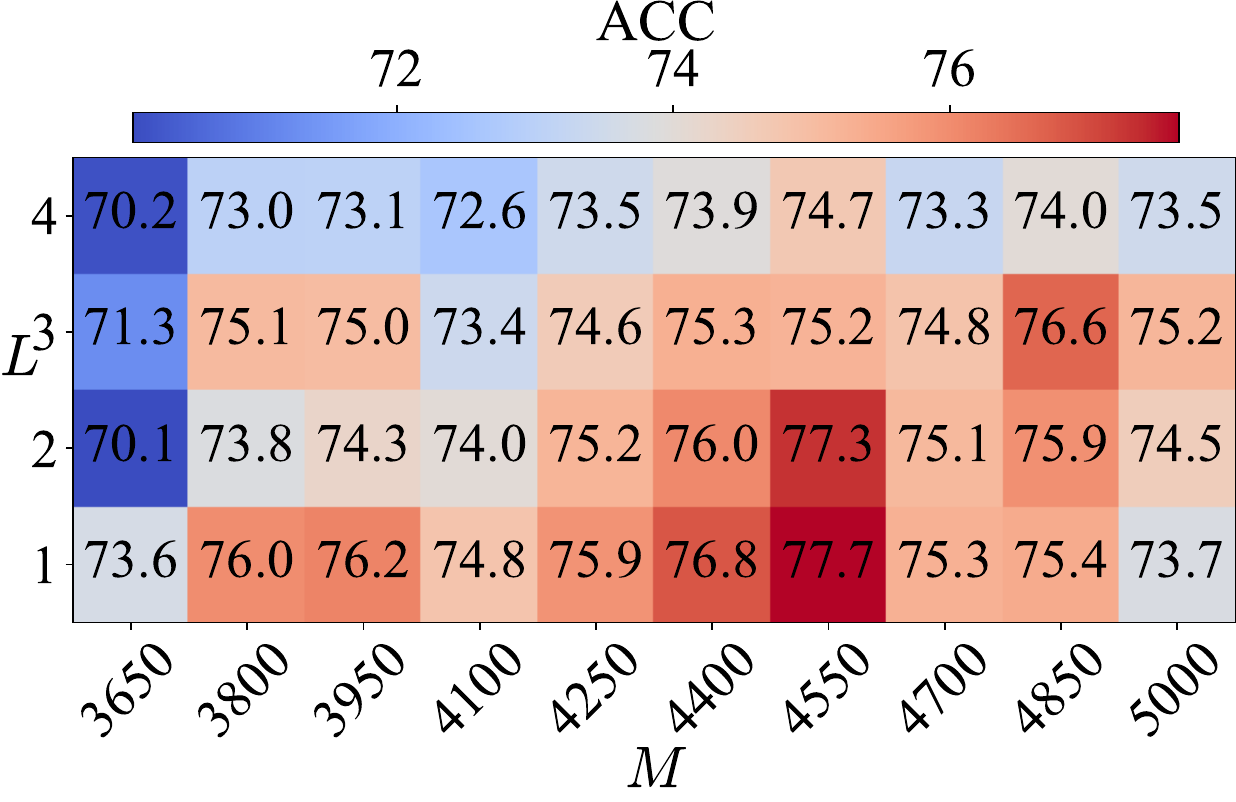}}
  \hfill
  \subfloat[]{\includegraphics[width=0.247\textwidth]{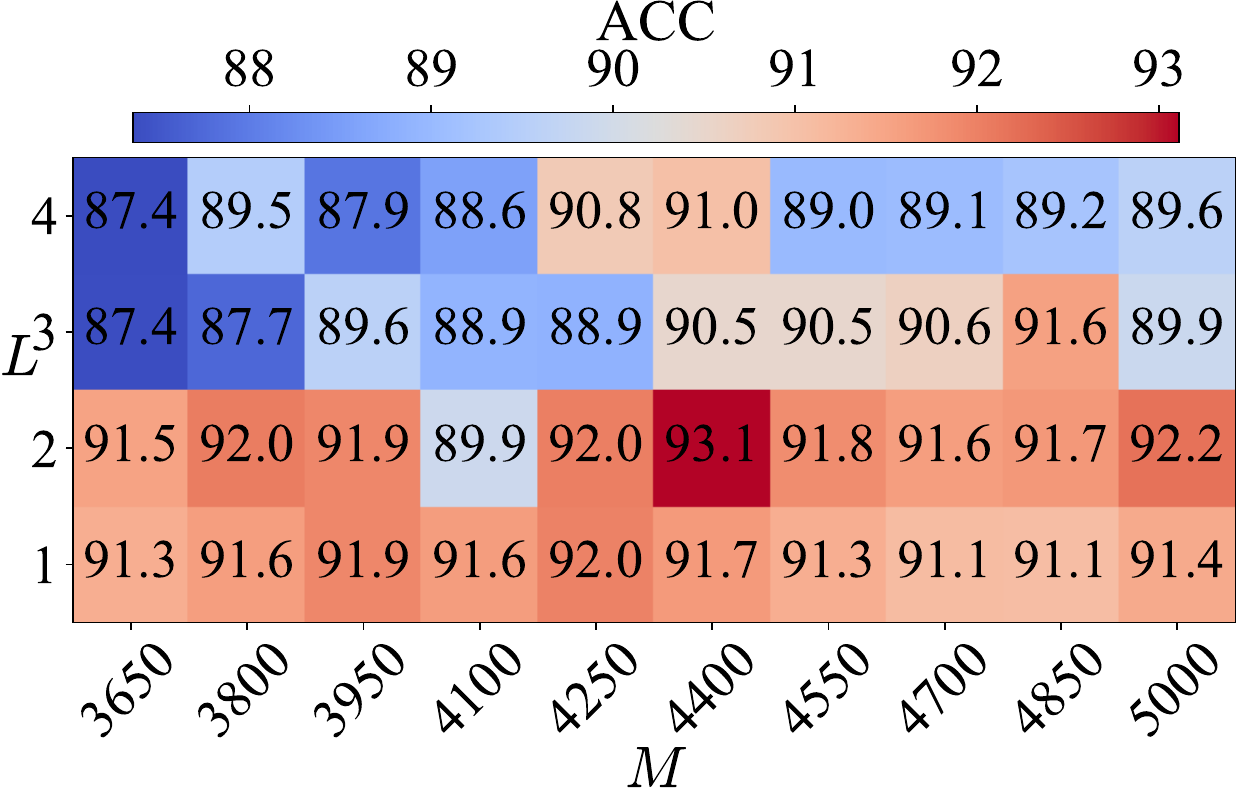}}
  \hfill
  \caption{The influence of different numbers of superpixels ($M$) and different numbers of network layers ($L$) on: (a) \textit{Indian Pines}, (b) \textit{Pavia University}, (c) \textit{Botswana} and (d) \textit{Trento}.}
  \label{nsp_nlayer}
\end{figure*}

\begin{figure*}[t]
  \centering
  \hfill
  \subfloat[]{\includegraphics[width=0.185\textwidth]{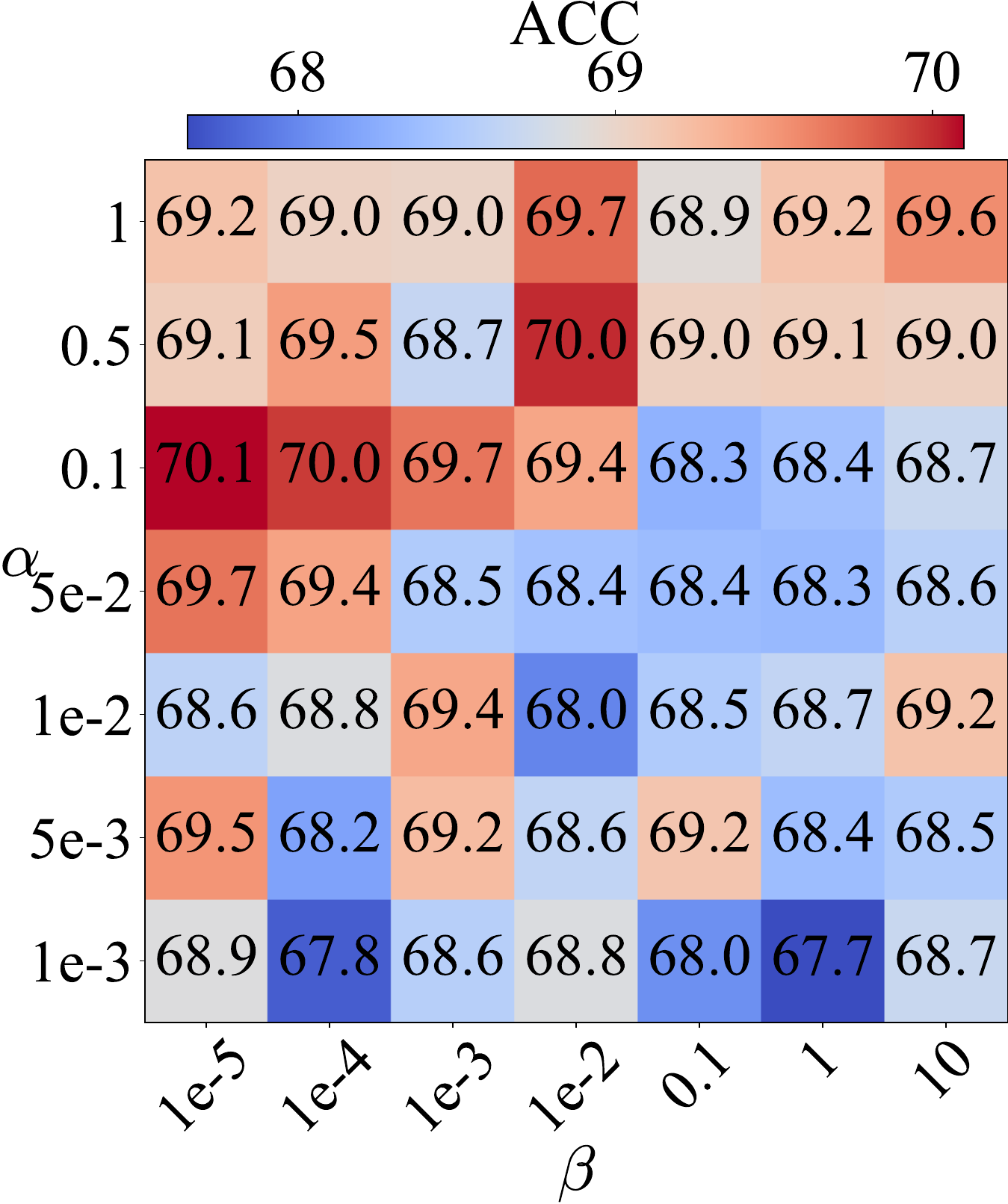}}
  \hfill
  \subfloat[]{\includegraphics[width=0.185\textwidth]{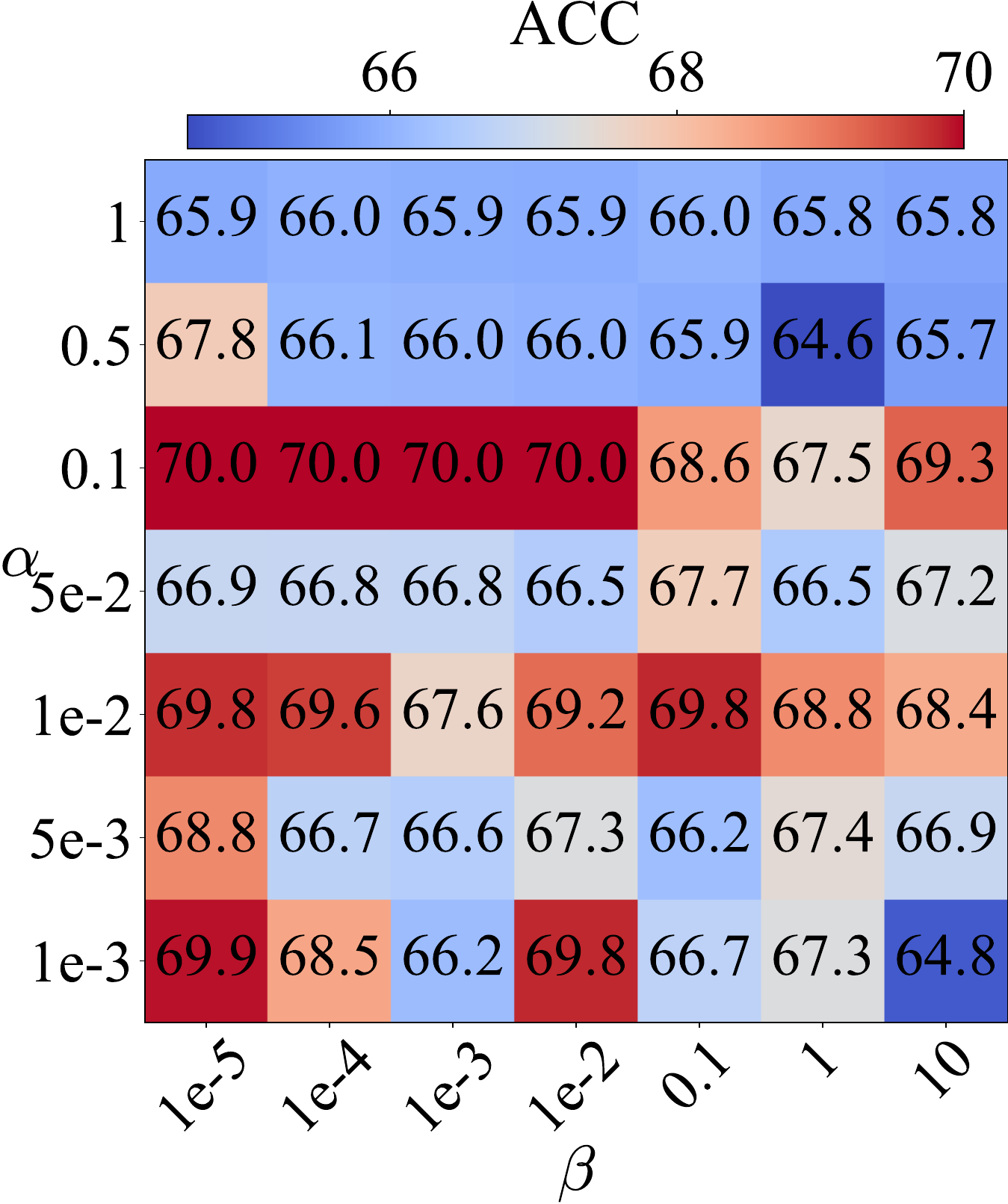}}
  \hfill
  \subfloat[]{\includegraphics[width=0.185\textwidth]{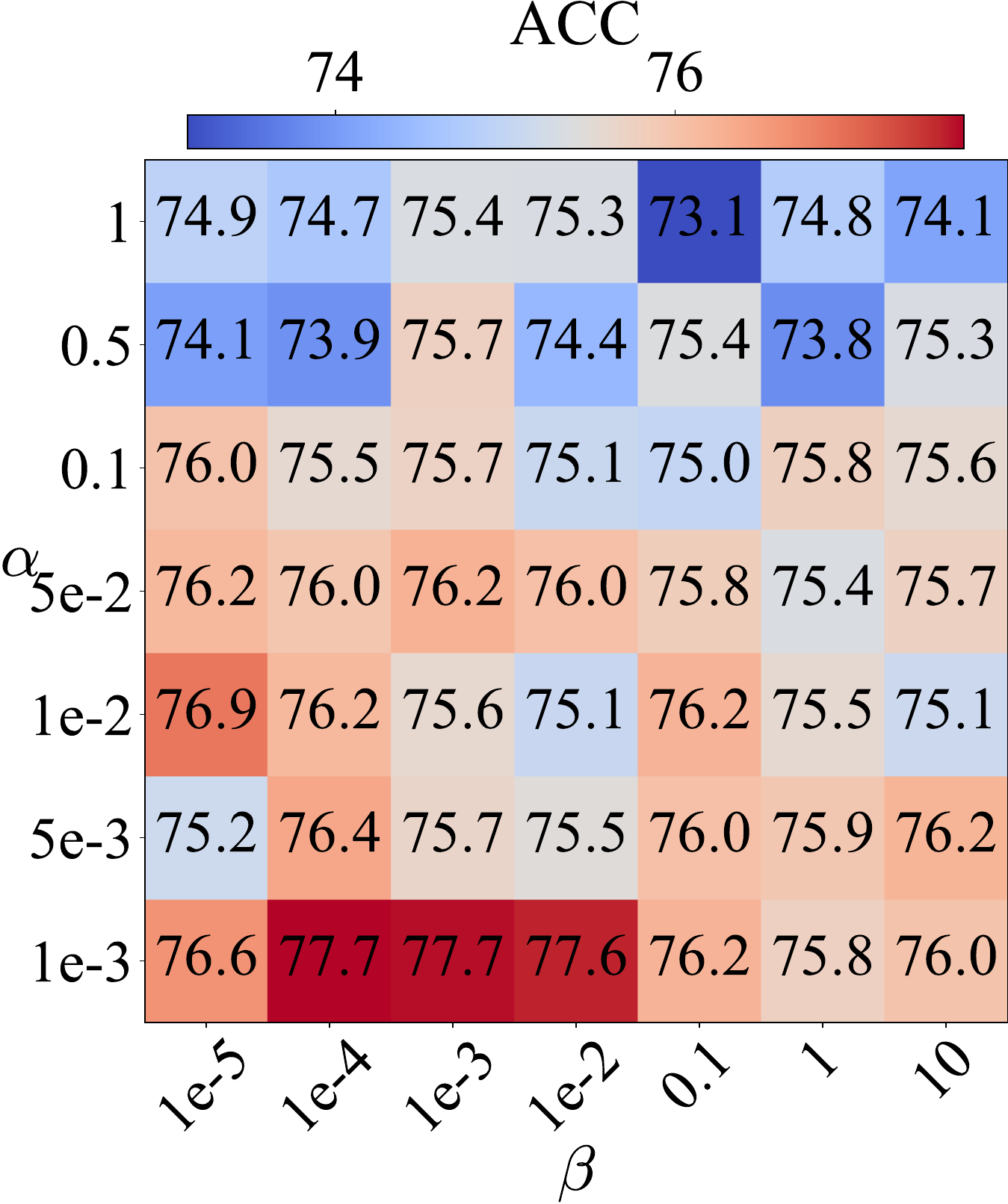}}
  \hfill
  \subfloat[]{\includegraphics[width=0.185\textwidth]{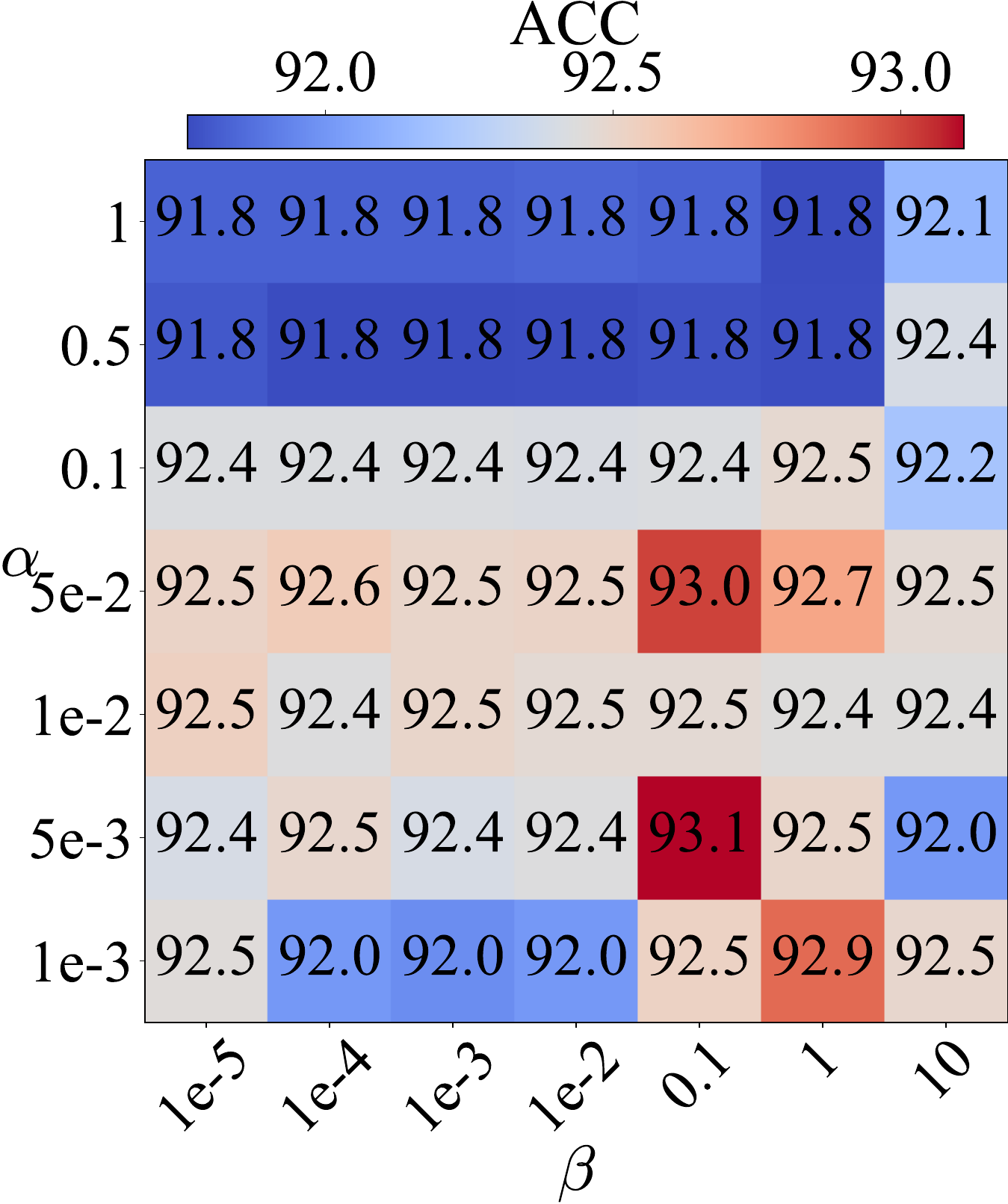}}
  \hfill
  \caption{The influence of different weights of $\mathcal{L}_{PC}$ ($\alpha$) and different weights of $\mathcal{L}_{E}$ ($\beta$) on: (a) \textit{Indian Pines}, (b) \textit{Pavia University}, (c) \textit{Botswana} and (d) \textit{Trento}.}
  \label{alpha_beta}
\end{figure*}

\begin{figure*}[t]
  \centering
  \hfill
  \subfloat[]{\includegraphics[width=0.245\textwidth]{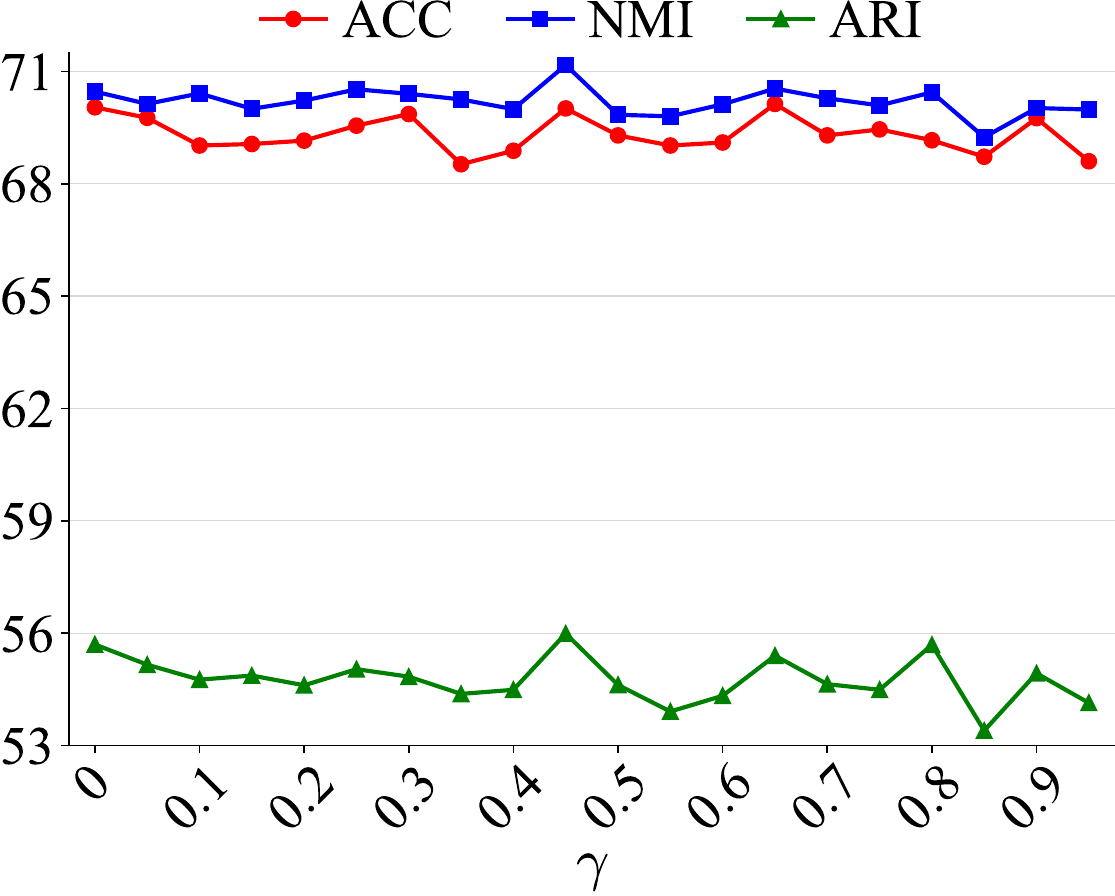}}
  \hfill
  \subfloat[]{\includegraphics[width=0.245\textwidth]{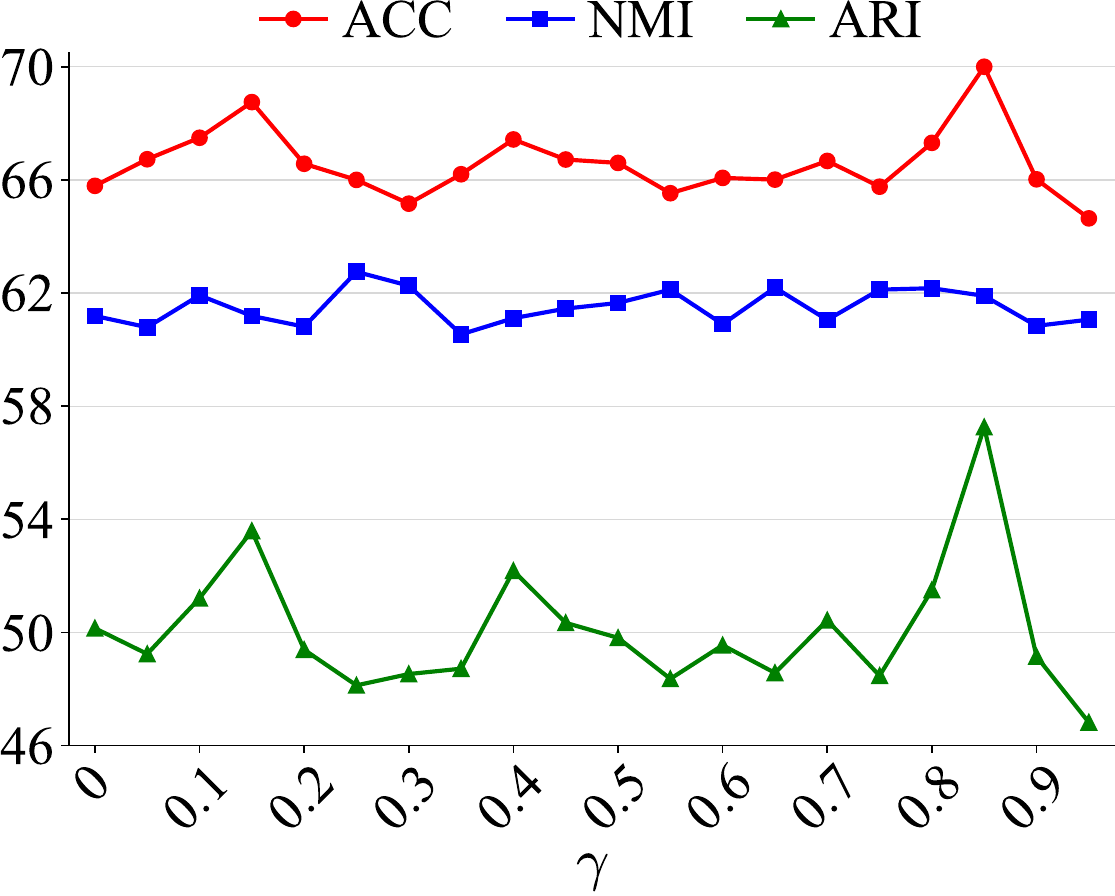}}
  \hfill
  \subfloat[]{\includegraphics[width=0.245\textwidth]{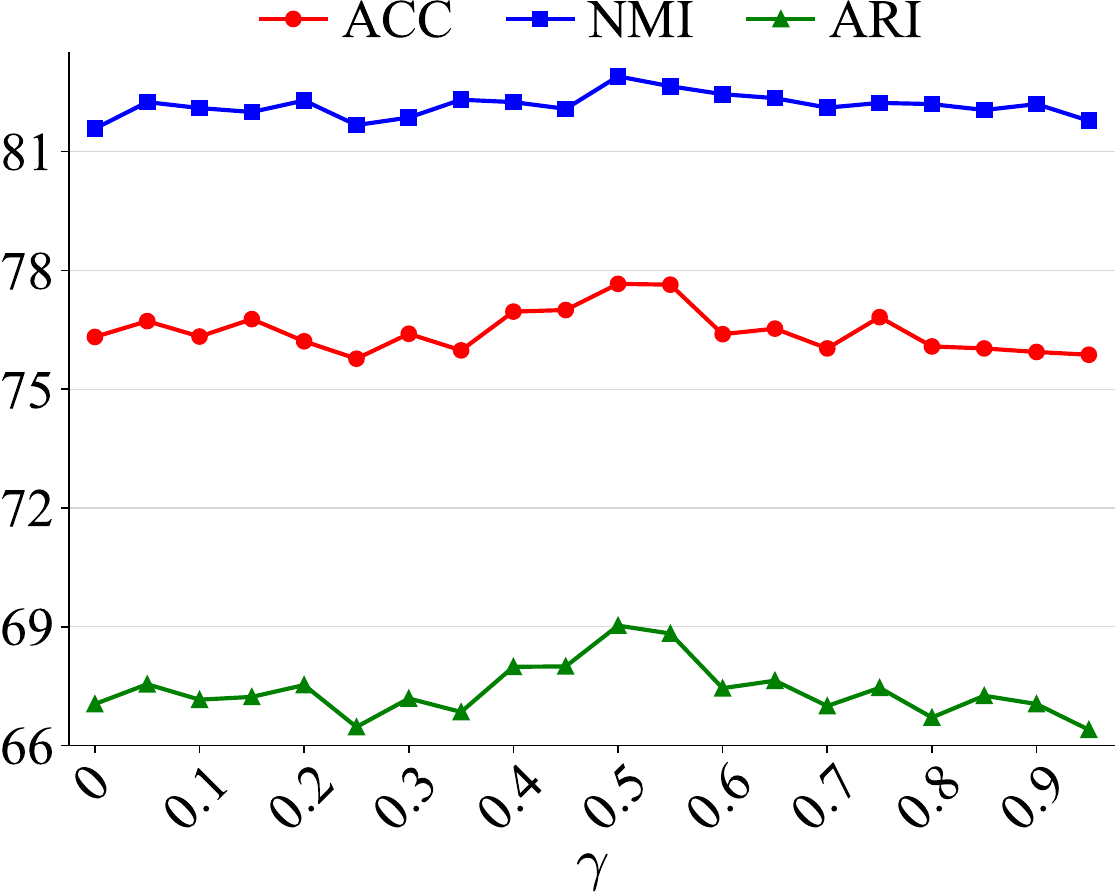}}
  \hfill
  \subfloat[]{\includegraphics[width=0.245\textwidth]{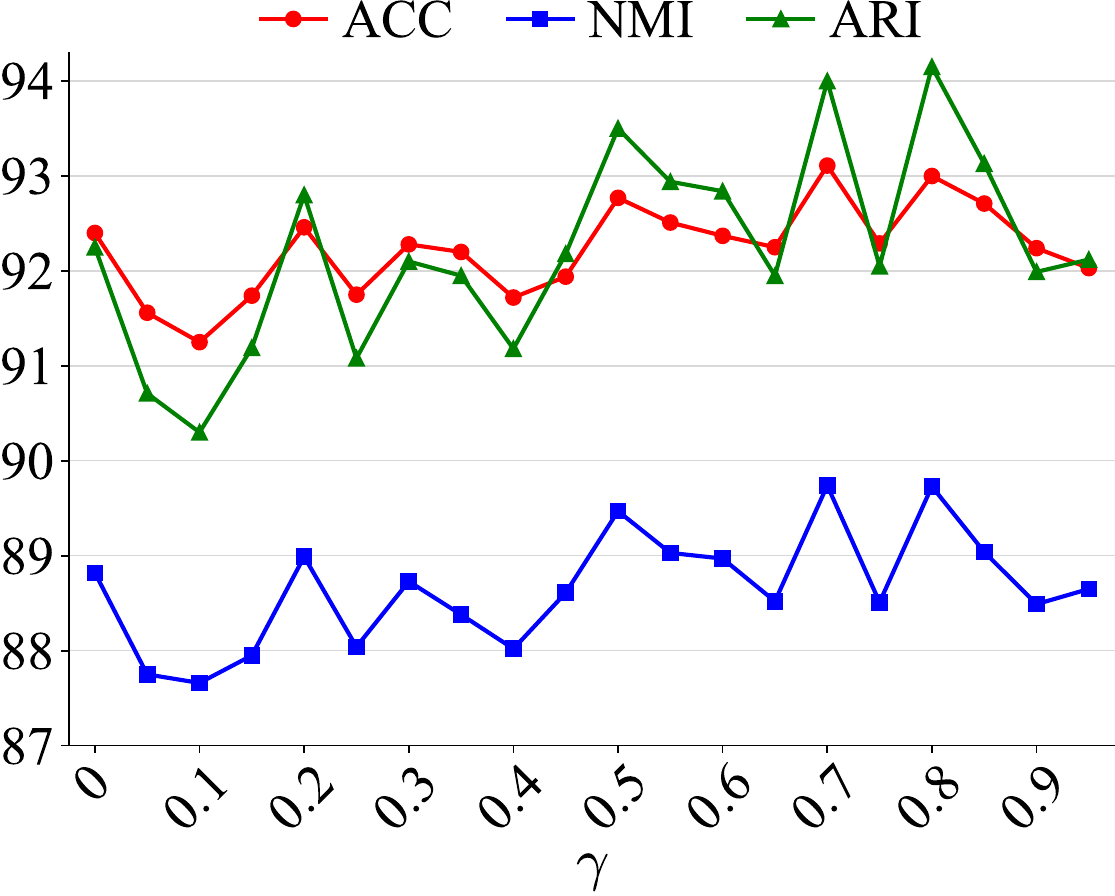}}
  \hfill
  \caption{The influence of different momentum coefficients of edge weight updating ($\gamma$) on: (a) \textit{Indian Pines}, (b) \textit{Pavia University}, (c) \textit{Botswana} and (d) \textit{Trento}.}
  \label{gamma_influence}
\end{figure*}

\subsection{Comparison with State-of-the-Art Methods}
We report the clustering performance of our method and other compared methods on \textit{Indian Pines}, \textit{Pavia University}, \textit{Botswana}, and \textit{Trento} in Table \ref{clustering_res}. For each metric, the best value is marked in bold, while the second-best value is underlined.

According to the quantitative clustering metrics, it could be observed that our method outperforms other compared methods by a large margin. For instance, our method improves the ACC by 2.61\%, 6.06\%, 4.96\% and 3.15\% on the four datasets. Notably, on \textit{Pavia University}, ACC, $\mathcal{K}$, ARI and F1 obtained by our method outperform the second-best value by 6.06\%, 7.47\%, 8.22\% and 1.13\%, respectively. Overall, on \textit{Botswana} and \textit{Trento}, our method achieves the best scores on all metrics; on \textit{Indian Pines}, it achieves the best scores on five metrics and the second-best on three metrics; on \textit{Pavia University}, it achieves the best scores on seven metrics and the second-best on one metric. Based on the results of pairwise $t$-Test at 0.05 significance level, our method significantly outperforms the compared methods in 91.8\% (323/352) cases.

Compared with subspace-clustering-based methods (i.e., SSC, NCSC and SGLSC) utilizing only spatial information, our method generally brings an improvement of more than 10\%. On the basis of the baseline method SSC, SGLSC enhances the similarity between adjacent superpixels by introducing a local superpixel similarity graph. This demonstrates the effectiveness of exploiting spatial information in HSI, which is also the objective behind the use of GCN in our method. Similarly, NCSC achieves the same objective by incorporating neighborhood contrastive regularization. Notably, NCSC leverages 2D convolution to enhance spatial feature extraction at the pixel level. However, the subsequent deep subspace clustering network requires a full batch input, thus incurring substantial memory cost. This limitation motivates us to propose the structural-spectral graph convolutional operator that conducts feature extraction entirely at the superpixel level, thereby improving computational efficiency.

Compared with the recent graph contrastive clustering method SPGCC, although both methods employ a similar contrastive clustering framework and superpixel data augmentations, the proposed structural-spectral graph convolutional operator infuses the capability of spectral feature extraction into GCN thus obtaining better superpixel representations than SPGCC, and improves the adjacency relationships between superpixels via evidence guided adaptive edge learning. As a result, our method outperforms SPGCC on all the four datasets.

Moreover, the clustering map visualization of different methods on \textit{Indian Pines} is reported in Fig. \ref{IP_clustering_map}. (Visualizations of other datasets are reported in Supplement I.) According to the visualizations, it is evident that our method demonstrates superior clustering for large-scale and continuously distributed land covers, e.g., green and yellow areas in \textit{Indian Pines}. This advantage primarily stems from our method performing superpixel-level clustering and adopting the edge learning to construct a more accurate adjacency matrix for superpixels.

\begin{table}[t]
  \centering
  \caption{Optimal Hyperparameters.\label{optimal_hyperparameters}}
  \renewcommand\arraystretch{1.1}
  \begin{tabular}{c!{\vrule width \lightrulewidth}ccccc} 
  \toprule
  Dataset                   & $M$  & $L$ & $\alpha$ & $\beta$ & $\gamma$  \\ 
  \midrule
  \textit{Indian Pines}     & 275  & 2   & 0.5      & 0.01    & 0.45      \\
  \textit{Pavia University} & 1000 & 4   & 0.1      & 0.001   & 0.85      \\
  \textit{Botswana}         & 4550 & 1   & 0.001    & 0.001   & 0.5       \\
  \textit{Trento}           & 4400 & 2   & 0.005    & 0.1     & 0.7       \\
  \bottomrule
  \end{tabular}
\end{table}

\begin{table}
\centering
\caption{Running Time of Different Methods (s), \textit{Indian Pines}, \textit{Pavia University}, \textit{Botswana} and \textit{Trento} are Abbreviated to \textit{IP}, \textit{PU}, \textit{BO} and \textit{TR}.\label{running_time}}
\renewcommand\arraystretch{1.11}
\begin{tabular}{c!{\vrule width \lightrulewidth}cccc!{\vrule width \lightrulewidth}c!{\vrule width \lightrulewidth}c} 
\toprule
Method     & \textit{IP} & \textit{PU} & \textit{BO} & \textit{TR} & Avg.   & Rk.  \\ 
\midrule
K-means    & 22.6        & 113.7       & 273.9       & 23.7        & 108.5  & 6    \\
SSC        & 20.3        & 16.4        & 22.3        & 9.7         & 17.2   & 1    \\
NCSC       & 77.5        & 6574.4      & 8559.3      & 2900.6      & 4528.0 & 12   \\
SGLSC      & 108.0       & 195.6       & 368.5       & 173.0       & 211.3  & 8    \\
DGAE       & 25.5        & 202.0       & 12.9        & 44.4        & 71.2   & 4    \\
EGRC       & 18.5        & 40.7        & 324.9       & 42.1        & 106.5  & 5    \\
S$^3$-ULDA & 11.5        & 374.5       & 2792.4      & 88.9        & 816.8  & 10   \\
S$^2$DL    & 12.5        & 544.9       & 2960.6      & 65.7        & 895.9  & 11   \\
SDST       & 296.9       & 229.7       & 169.0       & 129.2       & 206.2  & 7    \\
SPGCC      & 85.7        & 619.9       & 1129.2      & 437.3       & 568.0  & 9    \\
SAPC       & 11.1        & 21.8        & 51.6        & 13.2        & 24.4   & 2    \\
Ours       & 35.0        & 58.5        & 51.3        & 113.0       & 64.5   & 3    \\
\bottomrule
\end{tabular}
\end{table}

\subsection{Hyperparameters Analysis}
The proposed method mainly requires five hyperparameters: number of superpixels ($M$), number of network layers ($L$),  weight of $\mathcal{L}_{PC}$ ($\alpha$), weight of $\mathcal{L}_{E}$ ($\beta$) and momentum coefficient of edge weight updating ($\gamma$). We have tuned these hyperparameters for different datasets, and the adopted values are reported in Table \ref{optimal_hyperparameters}.

\textit{1) Impact of $M$ and $L$:} The number of superpixels ($M$) controls the segmentation of HSI and the homogeneity of superpixels, while the number of network layers ($L$) controls the margin of information aggregation and the receptive field of spectral feature convolution. For the proposed structural-spectral graph convolutional operator, one layer network enables superpixels to access their 1-hop neighbors. Large $M$ with small $L$ may lead to insufficient aggregation of spatial information in HSI, while small $M$ with large $L$ may result in over-smoothing representations of superpixels. As a result, according to Fig. \ref{nsp_nlayer}, the optimal $M$ on \textit{Indian Pines}, \textit{Pavia University}, \textit{Botswana} and \textit{Trento} are 275, 1000, 4550 and 4400, while the optimal $L$ are 2, 4, 1 and 2 respectively.

\textit{2) Impact of $\alpha$ and $\beta$:} $\alpha$ balances the influence between sample alignment and prototype contrast, while $\beta$ controls whether the predicted edge weights are closer to the empirical edge weights or influenced by contrastive clustering. As shown in Fig. \ref{alpha_beta}, our method is robust to $\alpha$ and $\beta$. We recommend selecting $\alpha$ and $\beta$ from \{0.01, 0.1\}, as these values consistently yield promising clustering performance on different datasets.

\textit{3) Impact of $\gamma$:} The momentum coefficient of edge weight updating ($\gamma$) determines the retention ratio of previous edge weights when updating the adjacency matrix. A smaller $\gamma$ indicates a more aggressive update strategy. Generally, the proposed method is also robust to $\gamma$. According to Fig. \ref{gamma_influence}, we suggest that $\gamma$ could be set between [0.5, 0.85].

\subsection{Running Time Analysis}
We report the running time of our method and other compared methods on four datasets in Table \ref{running_time}. In terms of the average running time, our method is only slower than SSC and SAPC, while faster than the other nine methods, demonstrating that our method is lightweight and efficient. Compared to SSC, our method achieves significantly better clustering performance on all datasets. Compared to SAPC, our method demonstrates an improvement of over 10\% in ACC on \textit{Indian Pines} and \textit{Botswana}, and an improvement of 4\%-6\% on other datasets. Therefore, the slight increase in running time is acceptable.

\begin{table*}[t]
  \centering
  \caption{Clustering Performance of Variant Networks.\label{variant_network_structure_results}}
  \resizebox{\textwidth}{!}{
  \renewcommand\arraystretch{1.14}
  \begin{tabular}{c!{\vrule width \lightrulewidth}ccc!{\vrule width \lightrulewidth}ccc!{\vrule width \lightrulewidth}ccc!{\vrule width \lightrulewidth}ccc} 
  \toprule
  \multirow{2}{*}{\raisebox{-8.5pt}{\begin{tabular}[c]{@{}c@{}}Network\\Variation\end{tabular}}} & \multicolumn{3}{c!{\vrule width \lightrulewidth}}{\textit{Indian Pines}} & \multicolumn{3}{c!{\vrule width \lightrulewidth}}{\textit{Pavia University}} & \multicolumn{3}{c!{\vrule width \lightrulewidth}}{\textit{Botswana}} & \multicolumn{3}{c}{\textit{Trento}}                              \\ 
  \cmidrule{2-13}
                                                                               & ACC                 & NMI                 & ARI                          & ACC                 & NMI                 & ARI                              & ACC                 & NMI                 & ARI                      & ACC                 & NMI                 & ARI                  \\ 
  \midrule
  MLP                                                                          & 54.77±1.73          & 56.89±1.69          & 37.05±2.31                   & 53.98±2.32          & 58.66±1.23          & 38.37±2.71                       & 66.08±3.06          & 71.84±2.62          & 53.61±3.39               & 75.92±1.36          & 74.08±1.20          & 68.14±1.57           \\
  1D-Conv                                                                      & 56.54±0.68          & 58.16±0.92          & 37.98±0.62                   & 60.88±3.01          & \textbf{65.31±2.85} & 47.12±6.27                       & 74.37±4.53          & 79.03±1.83          & 63.91±4.73               & 77.66±3.58          & 71.89±2.88          & 66.52±4.17           \\
  Graph-Conv                                                                   & 64.31±0.70          & 66.47±0.72          & 48.67±1.65                   & 56.62±2.69          & 47.23±2.15          & 40.53±6.57                       & 71.03±2.13          & 75.65±2.05          & 59.31±3.06               & 87.24±4.08          & 81.85±3.04          & 84.40±5.83           \\
  1D-Graph-Conv                                                                & 67.36±0.95          & 68.55±0.81          & 52.87±0.88                   & 63.72±2.21          & 60.20±1.45          & 46.09±1.47                       & 70.40±3.57          & 78.56±3.04          & 60.28±5.41               & 90.94±4.02          & 88.10±2.56          & 90.33±5.60           \\
  Graph-1D-Conv                                                                & 62.18±3.33          & 65.90±1.69          & 46.22±2.88                   & 51.73±5.83          & 46.59±5.26          & 36.44±8.49                       & 66.08±2.53          & 74.79±1.58          & 55.69±1.34               & 92.06±0.97          & 87.01±0.45          & 92.49±0.83           \\
  SSGCO                                                                        & \textbf{70.01±1.95} & \textbf{71.17±1.31} & \textbf{55.99±2.07}          & \textbf{70.00±3.46} & 61.90±3.18          & \textbf{57.26±8.57}              & \textbf{77.66±1.28} & \textbf{82.90±1.30} & \textbf{69.03±2.38}      & \textbf{93.11±0.96} & \textbf{89.74±0.94} & \textbf{94.00±1.30}  \\
  \bottomrule
  \end{tabular}}
\end{table*}

\begin{table*}[t]
\centering
\caption{Ablation Study Results of Evidence Guided Adaptive Edge Learning.\label{ablation_study_results_EGAEL}}
\resizebox{\textwidth}{!}{
\renewcommand\arraystretch{1.14}
\begin{tabular}{cc!{\vrule width \lightrulewidth}ccc!{\vrule width \lightrulewidth}ccc!{\vrule width \lightrulewidth}ccc!{\vrule width \lightrulewidth}ccc} 
\toprule
\multicolumn{2}{c!{\vrule width \lightrulewidth}}{Item} & \multicolumn{3}{c!{\vrule width \lightrulewidth}}{\textit{Indian Pines}} & \multicolumn{3}{c!{\vrule width \lightrulewidth}}{\textit{Pavia University}} & \multicolumn{3}{c!{\vrule width \lightrulewidth}}{\textit{Botswana}} & \multicolumn{3}{c}{\textit{Trento}}                              \\ 
\midrule
$w^{pre}$      & $w^{emp}$                                   & ACC                 & NMI                 & ARI                          & ACC                 & NMI                 & ARI                              & ACC                 & NMI                 & ARI                      & ACC                 & NMI                 & ARI                  \\ 
\midrule
             &                                           & 67.45±1.43          & 68.07±1.42          & 53.13±1.04                   & 64.96±1.40          & 60.54±4.22          & 46.20±4.27                       & 74.06±0.82          & 81.78±0.92          & 66.89±1.23               & 91.85±1.08          & 88.63±1.25          & 92.68±2.13           \\
$\checkmark$ &                                           & 69.54±1.90          & 70.02±1.30          & 54.79±2.91                   & 68.26±3.36          & \textbf{62.66±3.44} & 54.64±7.43                       & 76.64±1.97          & 82.32±0.92          & 67.93±2.28               & 92.40±1.53          & 89.05±1.42          & 92.86±2.51           \\
             & $\checkmark$                              & 67.86±0.72          & 68.88±1.05          & 52.81±1.23                   & 64.88±1.78          & 62.36±1.54          & 47.71±1.82                       & 73.39±2.14          & 81.10±1.14          & 65.50±1.75               & 90.90±1.46          & 87.68±1.49          & 90.99±2.94           \\
$\checkmark$ & $\checkmark$                              & \textbf{70.01±1.95} & \textbf{71.17±1.31} & \textbf{55.99±2.07}          & \textbf{70.00±3.46} & 61.90±3.18          & \textbf{57.26±8.57}              & \textbf{77.66±1.28} & \textbf{82.90±1.30} & \textbf{69.03±2.38}      & \textbf{93.11±0.96} & \textbf{89.74±0.94} & \textbf{94.00±1.30}  \\
\bottomrule
\end{tabular}}
\end{table*}

\begin{figure*}[t]
  \centering
  \hfill
  \subfloat[]{\includegraphics[width=0.245\textwidth]{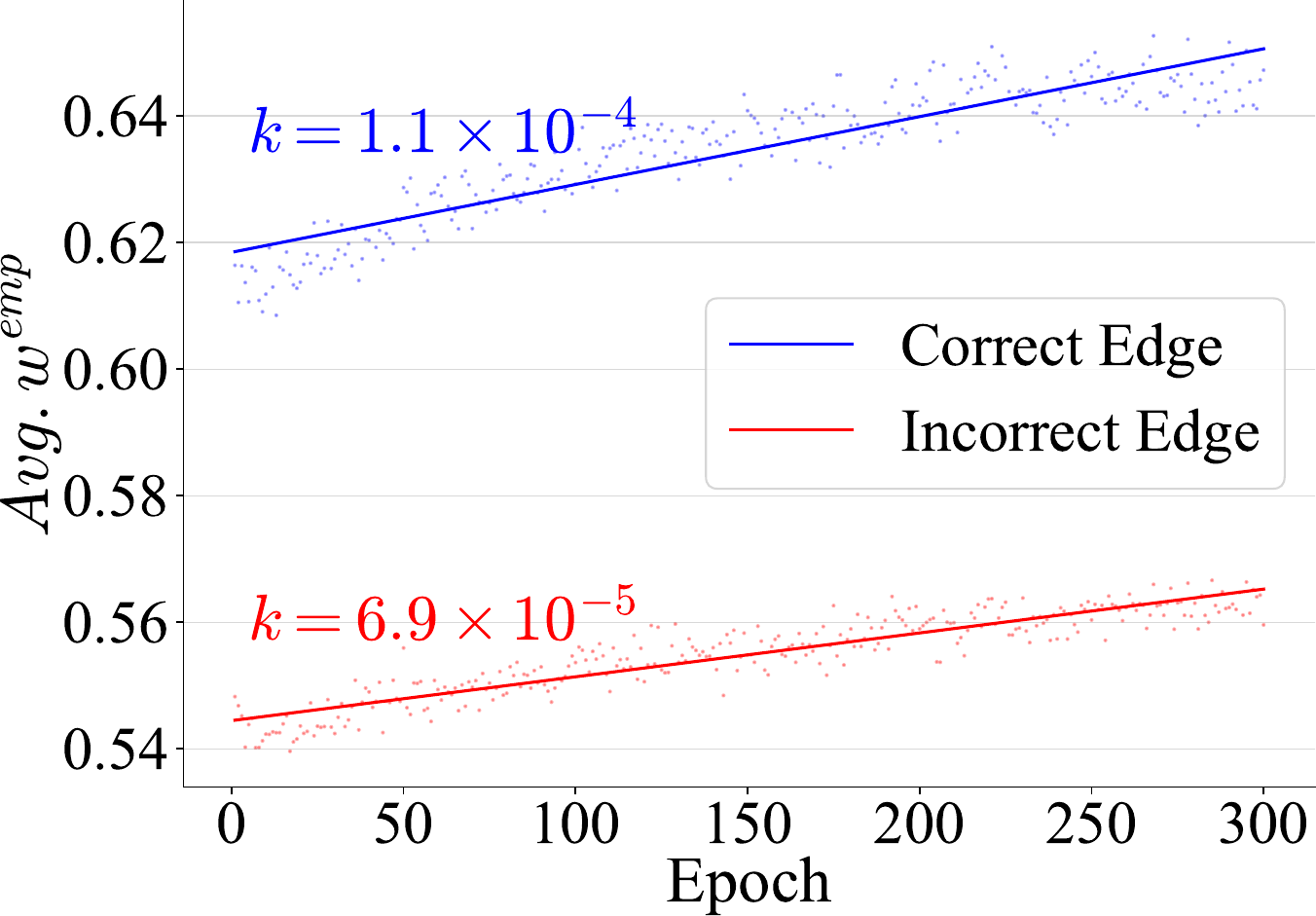}}
  \hfill
  \subfloat[]{\includegraphics[width=0.245\textwidth]{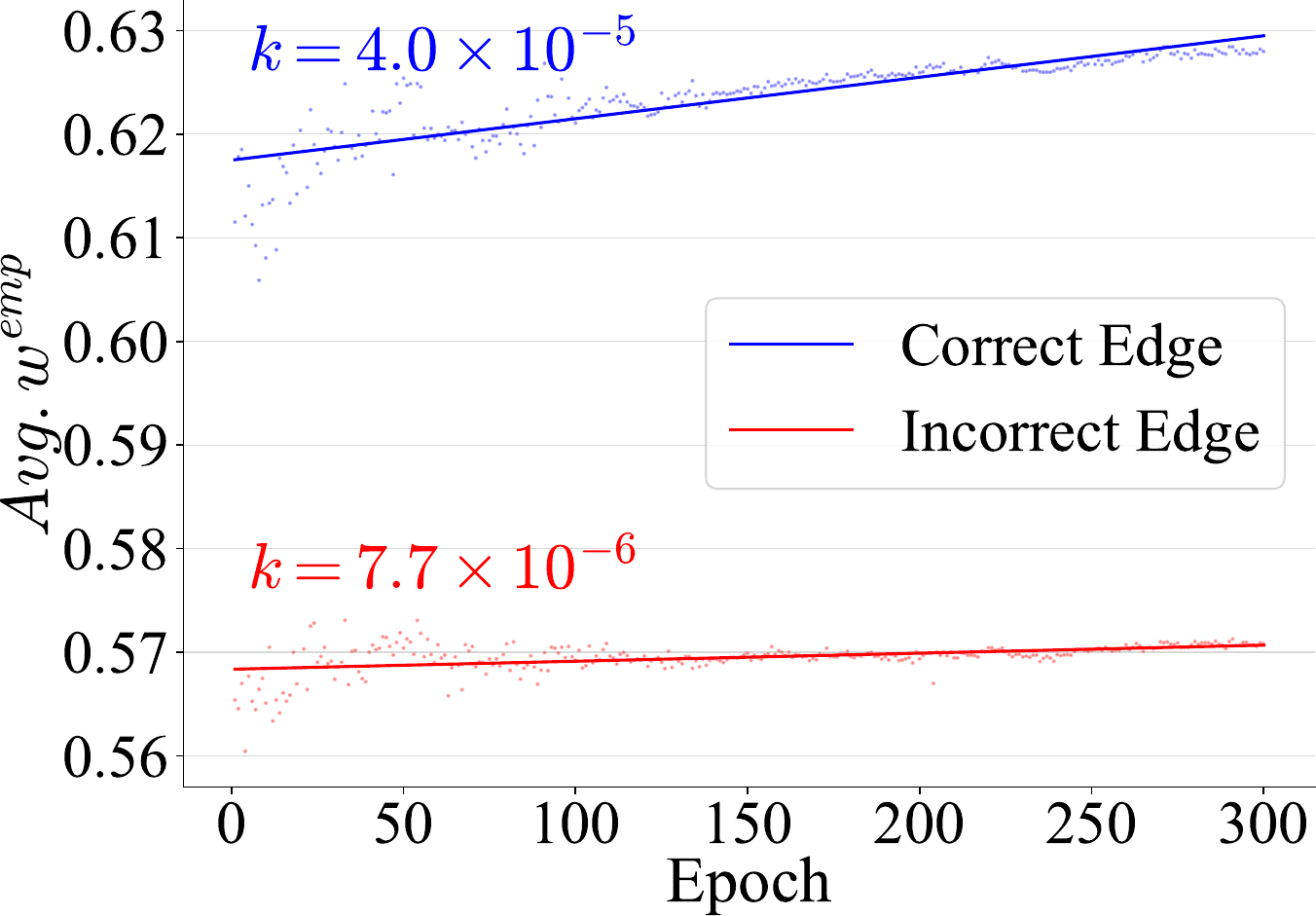}}
  \hfill
  \subfloat[]{\includegraphics[width=0.245\textwidth]{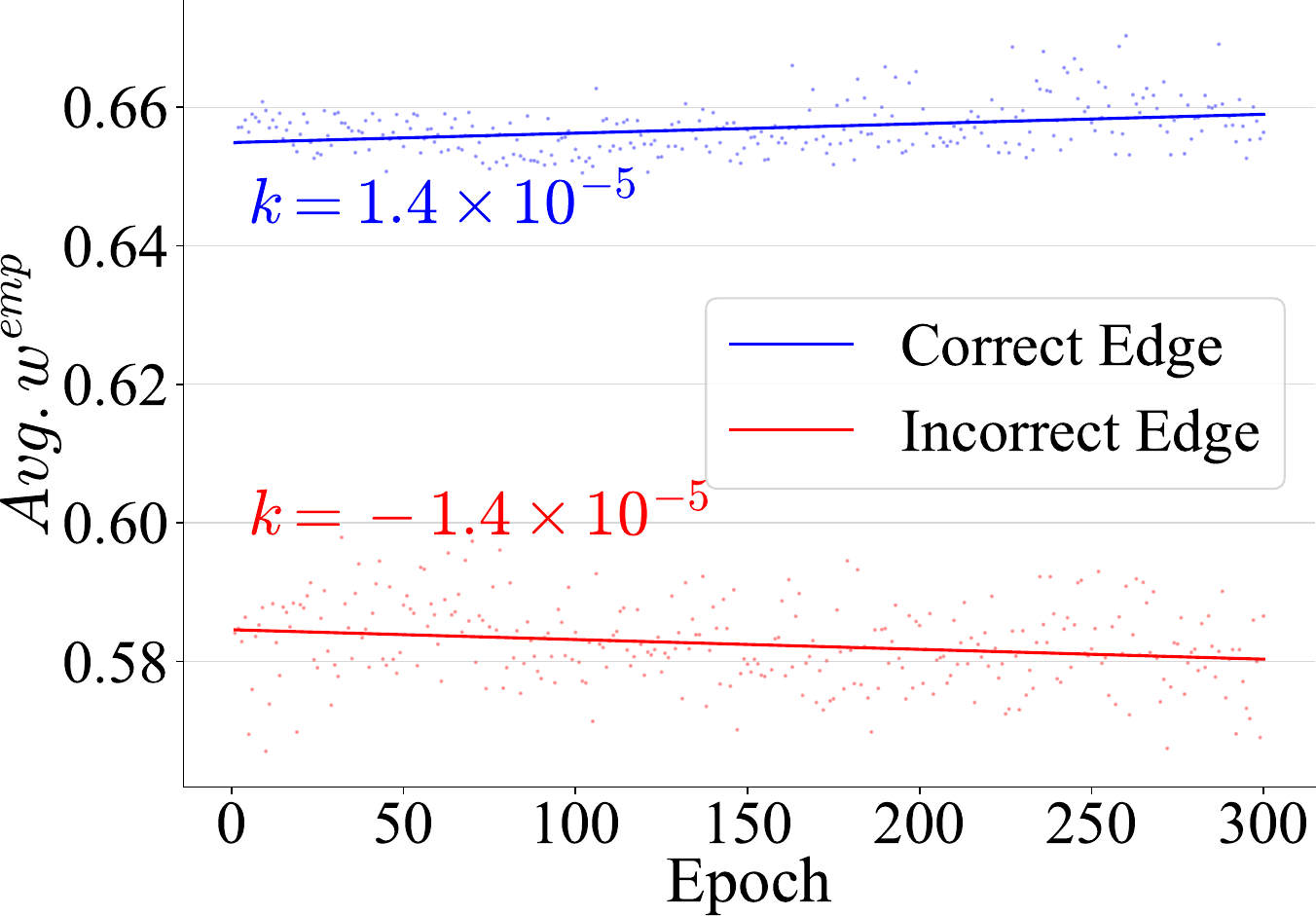}}
  \hfill
  \subfloat[]{\includegraphics[width=0.245\textwidth]{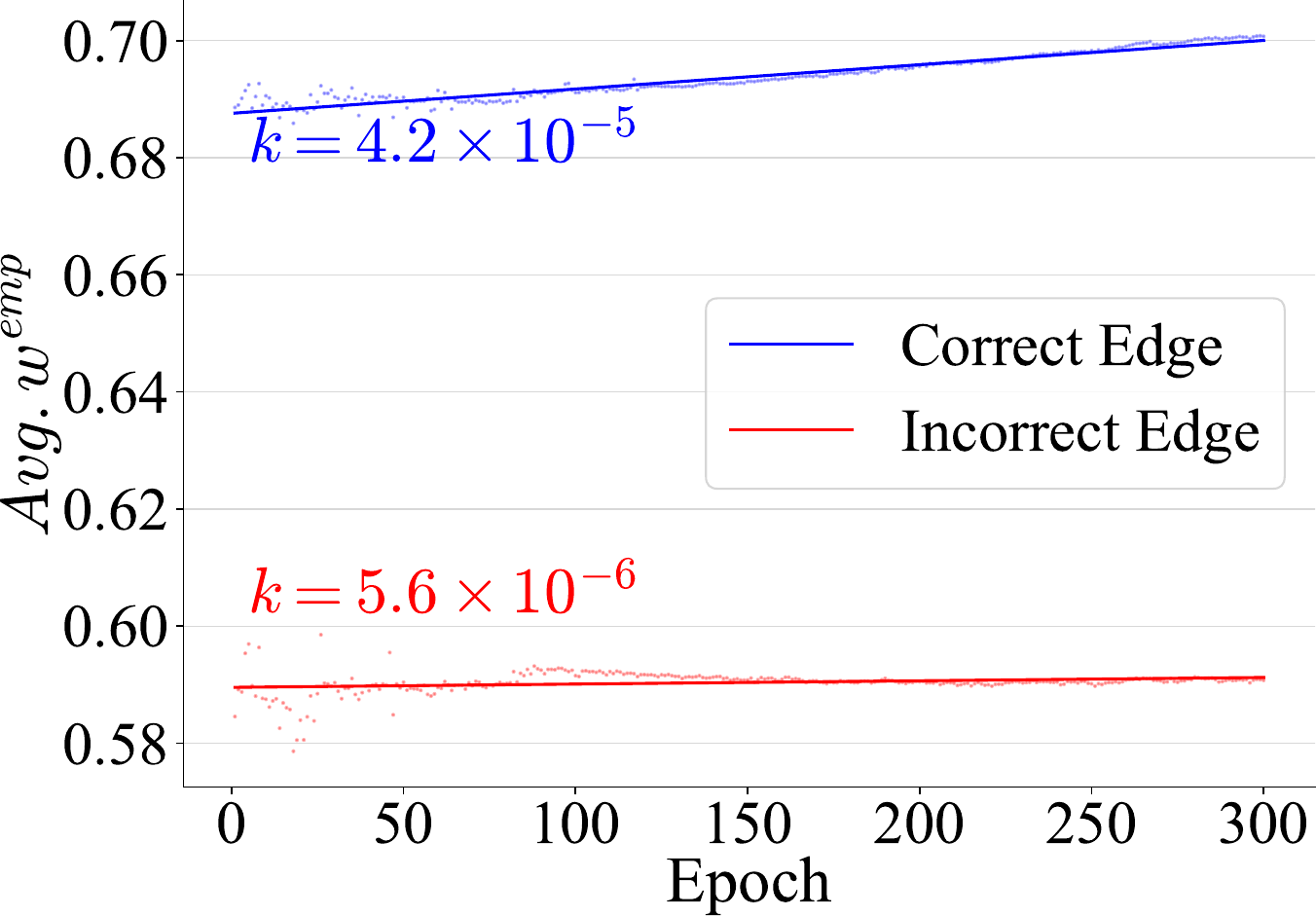}}
  \hfill
  \caption{Average empirical edge weight ($w^{emp}$) per epoch and the result of linear fitting on: (a) \textit{Indian Pines}, (b) \textit{Pavia University}, (c) \textit{Botswana} and (d) \textit{Trento}.}
  \label{w_emp_epoch}
\end{figure*}

\subsection{Further Analysis of Structural-Spectral Graph Convolutional Operator}
To validate the effectiveness of the proposed structural-spectral graph convolutional operator (SSGCO), we compared it with a series of variant superpixel representation networks, including MLP, 1D-Conv, Graph-Conv, 1D-Graph-Conv and Graph-1D-Conv, as detailed below:

\textit{MLP:} MLP that performs projection transformations is employed for superpixels.

\textit{1D-Conv:} Several 1D convolutional layers are used to extract spectral features.

\textit{Graph-Conv:} Several GCN layers are utilized to aggregate spatial information.

\textit{1D-Graph-Conv:} Several 1D convolutional layers are applied to extract spectral features, followed by several GCN layers to aggregate spatial information.

\textit{Graph-1D-Conv:} Several GCN layers are applied to aggregate spatial information, followed by several 1D convolutional layers to extract spectral features.

According to the experimental results presented in Table \ref{variant_network_structure_results}, the proposed SSGCO outperforms all other variant networks with a large margin. In addition, we have identified two observations.

First, compared to the MLP baseline, employing 1D-Conv alone improves the average ACC by 4.68\% on four datasets, while employing Graph-Conv alone improves the average ACC by 7.11\%. This demonstrates their capabilities of extracting high-order spectral features and aggregating spatial information. SSGCO integrates both spatial and spectral information of HSI effectively, thus leading to superior superpixel representations and clustering performance.

Second, comparing 1D-Graph-Conv, Graph-1D-Conv and SSGCO, both spectral feature extraction and spatial information aggregation are fused within each layer in SSGCO, while separated in 1D-Graph-Conv and Graph-1D-Conv. As a result, 1D-Graph-Conv and Graph-1D-Conv perform worse than SSGCO. Notably, Graph-1D-Conv yields the least favorable results. One possible reason is that several layers of projective transformations in GCN disrupt the continuity of spectral features, thereby limiting the effectiveness of subsequent 1D convolution.

\subsection{Further Analysis of Evidence Guided Adaptive Edge Learning}
In the proposed evidence guided adaptive edge learning (EGAEL) module, the predicted edge weight $w^{pre}$ is generated by a learnable MLP based on current node representations and used to update the adjacency matrix. Differently, the empirical edge weight $w^{emp}$ is calculated based on heuristic measures such as node similarity and clustering confidence. $w^{emp}$ does not directly modify the graph, but instead serves as supervision to guide the learning of $w^{pre}$, encouraging $w^{pre}$ to reflect more reliable inter-superpixel relationships.

To validate the effectiveness of $w^{pre}$ and $w^{emp}$, we first conducted an ablation study. As shown in Table \ref{ablation_study_results_EGAEL}, the complete EGAEL module results in an average ACC improvement of 3.12\%, while leveraging either $w^{pre}$ or $w^{emp}$ individually yields inferior performance, underscoring the importance of their interaction within EGAEL module.

Next, we analyzed the value changes of $w^{emp}$ during the training. We classified edges as correct or incorrect based on ground truth superpixel labels, which are determined by the majority class of internal labeled pixels. A correct edge connects two superpixels of the same class, otherwise, it is incorrect. We computed the average $w^{emp}$ of correct and incorrect edges at each epoch, and conducted linear fitting, as shown in Fig. \ref{w_emp_epoch}. On all datasets, the average $w^{emp}$ of correct edges is consistently higher than that of incorrect edges. According to the fitted lines, correct edges exhibit significantly larger slope $k$ than incorrect edges, indicating an increasing gap between correct and incorrect edges.

Then, we interpreted $w^{pre}$ as the probability of an edge being correct, and determined an optimal threshold to classify correct and incorrect edges. The classification performance is compared against that of using initial edge weights $w^{init}$, which assume all edges are correct. As shown in Table \ref{w_pre_acc}, $w^{pre}$ achieves higher accuracy on all the four datasets, with an improvement of 6.64\% on \textit{Indian Pines} dataset in particular. This demonstrates that the adjacency matrix constructed from $w^{pre}$ is more reliable than the initial one, thereby validating the effectiveness of the proposed EGAEL module.

\begin{table}[t]
  \centering
  \caption{Classification Accuracy of Correct and Incorrect Edges with Initial Edge Weights ($w^{init}$) and Predicted Edge Weights ($w^{pre}$).\label{w_pre_acc}}
  \renewcommand\arraystretch{1.1}
  \begin{tabular}{c!{\vrule width \lightrulewidth}cccc} 
  \toprule
  Edge Type      & \textit{Indian Pines} & \textit{Pavia University} & \textit{Botswana} & \textit{Trento}  \\ 
  \midrule
  $w^{init}$       & 62.33\%                 & 69.89\%                     & 88.69\%             & 97.40\%            \\
  $w^{pre}$        & 68.97\%                 & 73.54\%                     & 90.59\%             & 97.64\%            \\ 
  \midrule
  Inc.$\uparrow$ & 6.64\%                  & 3.65\%                      & 1.90\%              & 0.24\%             \\
  \bottomrule
  \end{tabular}
\end{table}

\section{Conclusion\label{conclusion}}
This paper presents a superpixel clustering method for HSI, centered on two key innovations: a structural-spectral graph convolutional operator (SSGCO) for enhanced HSI superpixel representation and an evidence guided adaptive edge learning (EGAEL) module for superpixel topological graph refinement. SSGCO infuses dedicated spectral convolution into the information aggregation of GNNs, enabling comprehensive extraction of both spatial and spectral information from superpixels. Complementarily, EGAEL adaptively optimizes the superpixel topological graph by learning edge weights under the guidance of empirical edge weights, effectively mitigating the influence of initial incorrect connections. Finally, we integrate the proposed method into a contrastive clustering framework, leveraging sample alignment and prototype contrast as clustering-oriented training objectives to achieve clustering. Extensive experiments on several HSI datasets demonstrate the effectiveness and superiority of our method.

Nevertheless, the proposed method still has some limitations. At the beginning of our method, we represent each superpixel as the average of its internal pixel features. This common practice, however, leads to the loss of 2D spatial structure within superpixels. In future work, we aim to exploit superpixels' internal spatial information. For example, we will consider padding superpixels into regular rectangles and introducing 2D convolution to learn better representations.

\bibliographystyle{IEEEtran}
\bibliography{main_bib}

@article{app_precision_agriculture,
  title   = {A systematic review of hyperspectral imaging in precision agriculture: Analysis of its current state and future prospects},
  journal = {Comput. Electron. Agric.},
  volume  = {222},
  pages   = {109037},
  year    = {2024},
  issn    = {0168-1699},
  author  = {Billy G. Ram and Peter Oduor and C. Igathinathane and Kirk Howatt and Xin Sun}
}

@article{hsi_cls_1,
  author  = {Liu, Hui and Jia, Yuheng and Hou, Junhui and Zhang, Qingfu},
  journal = {IEEE Trans. Circuits Syst. Video Technol.},
  title   = {Global-Local Balanced Low-Rank Approximation of Hyperspectral Images for Classification},
  year    = {2022},
  volume  = {32},
  number  = {4},
  pages   = {2013-2024},
  doi     = {10.1109/TCSVT.2021.3095250}
}

@ARTICLE{hsi_cls_2,
  author={Han, Bo and Jia, Yuheng and Liu, Hui and Hou, Junhui},
  journal={IEEE Trans. Image Process.}, 
  title={Irregular Tensor Low-Rank Representation for Hyperspectral Image Representation}, 
  year={2025},
  volume={34},
  number={},
  pages={3239-3252},
  doi={10.1109/TIP.2025.3571669}}

@ARTICLE{hsi_cls_3,
  author={Hong, Danfeng and Gao, Lianru and Yao, Jing and Zhang, Bing and Plaza, Antonio and Chanussot, Jocelyn},
  journal={IEEE Trans. Geosci. Remote Sens.}, 
  title={Graph Convolutional Networks for Hyperspectral Image Classification}, 
  year={2021},
  volume={59},
  number={7},
  pages={5966-5978},
  doi={10.1109/TGRS.2020.3015157}}

@ARTICLE{hsi_cls_4,
  author={Chen, Jianqi and Chen, Keyan and Chen, Hao and Li, Wenyuan and Zou, Zhengxia and Shi, Zhenwei},
  journal={IEEE Trans. Geosci. Remote Sens.}, 
  title={Contrastive Learning for Fine-Grained Ship Classification in Remote Sensing Images}, 
  year={2022},
  volume={60},
  number={},
  pages={1-16},
  doi={10.1109/TGRS.2022.3192256}}

@ARTICLE{hsi_cls_5,
  author={Yang, Shujun and Hou, Junhui and Jia, Yuheng and Mei, Shaohui and Du, Qian},
  journal={IEEE Trans. Image Process.}, 
  title={Superpixel-Guided Discriminative Low-Rank Representation of Hyperspectral Images for Classification}, 
  year={2021},
  volume={30},
  number={},
  pages={8823-8835},
  doi={10.1109/TIP.2021.3120675}}

@ARTICLE{hsi_cls_6,
  author={Hong, Danfeng and Han, Zhu and Yao, Jing and Gao, Lianru and Zhang, Bing and Plaza, Antonio and Chanussot, Jocelyn},
  journal={IEEE Trans. Geosci. Remote Sens.}, 
  title={SpectralFormer: Rethinking Hyperspectral Image Classification With Transformers}, 
  year={2022},
  volume={60},
  number={},
  pages={1-15},
  doi={10.1109/TGRS.2021.3130716}}

@ARTICLE{hsi_cls_7,
  author={Mei, Shaohui and Hou, Junhui and Chen, Jie and Chau, Lap-Pui and Du, Qian},
  journal={IEEE Trans. Geosci. Remote Sens.}, 
  title={Simultaneous Spatial and Spectral Low-Rank Representation of Hyperspectral Images for Classification}, 
  year={2018},
  volume={56},
  number={5},
  pages={2872-2886},
  doi={10.1109/TGRS.2017.2785359}}

@article{app_environmental_monitoring,
  author         = {Tang, Yang and Song, Shuang and Gui, Shengxi and Chao, Weilun and Cheng, Chinmin and Qin, Rongjun},
  title          = {Active and Low-Cost Hyperspectral Imaging for the Spectral Analysis of a Low-Light Environment},
  journal        = {Sensors},
  volume         = {23},
  year           = {2023},
  number         = {3},
  article-number = {1437},
  pubmedid       = {36772477},
  issn           = {1424-8220},
  doi            = {10.3390/s23031437}
}

@article{app_mineral_exploration,
  title   = {A review on hyperspectral imagery application for lithological mapping and mineral prospecting: Machine learning techniques and future prospects},
  journal = {Remote Sens. Appl.: Soc. Environ.},
  volume  = {35},
  pages   = {101218},
  year    = {2024},
  issn    = {2352-9385},
  author  = {Soufiane Hajaj and Abderrazak {El Harti} and Amin Beiranvand Pour and Amine Jellouli and Zakaria Adiri and Mazlan Hashim}
}

@article{app_urban_planning,
  author         = {Ma, Xiaotong and Man, Qixia and Yang, Xinming and Dong, Pinliang and Yang, Zelong and Wu, Jingru and Liu, Chunhui},
  title          = {Urban Feature Extraction within a Complex Urban Area with an Improved 3D-CNN Using Airborne Hyperspectral Data},
  journal        = {Remote Sens.},
  volume         = {15},
  year           = {2023},
  number         = {4},
  article-number = {992},
  issn           = {2072-4292},
  doi            = {10.3390/rs15040992}
}

@ARTICLE{spectral_noise,
  author={Li, Zhixin and Han, Bo and Jia, Yuheng},
  journal = {IEEE Geosci. Remote Sens. Lett.},
  title={Discriminative Low-Rank Representation for HSI Clustering}, 
  year={2024},
  volume={21},
  number={},
  pages={1-5},
  doi={10.1109/LGRS.2024.3465498}}

@article{kmeans,
  author  = {Kanungo, T. and Mount, D.M. and Netanyahu, N.S. and Piatko, C.D. and Silverman, R. and Wu, A.Y.},
  journal = {IEEE Trans. Pattern Anal. Mach. Intell.},
  title   = {An efficient k-means clustering algorithm: analysis and implementation},
  year    = {2002},
  volume  = {24},
  number  = {7},
  pages   = {881-892}
}

@inproceedings{fuzzycmeans,
  author    = {Yang, Tai-ning and Lee, Chih-jen and Yen, Shi-jim},
  booktitle = {Proc. IEEE Int. Conf. Fuzzy Syst. (FUZZ-IEEE)},
  title     = {Fuzzy objective functions for robust pattern recognition},
  year      = {2009},
  volume    = {},
  number    = {},
  pages     = {2057-2062}
}

@article{ssc,
  author  = {Elhamifar, Ehsan and Vidal, René},
  journal = {IEEE Trans. Pattern Anal. Mach. Intell.},
  title   = {Sparse Subspace Clustering: Algorithm, Theory, and Applications},
  year    = {2013},
  volume  = {35},
  number  = {11},
  pages   = {2765-2781}
}

@inproceedings{ers,
  author    = {Liu, Ming-Yu and Tuzel, Oncel and Ramalingam, Srikumar and Chellappa, Rama},
  booktitle = {Proc. IEEE/CVF Conf. Comput. Vis. Pattern Recognit. (CVPR)},
  title     = {Entropy rate superpixel segmentation},
  year      = {2011},
  volume    = {},
  number    = {},
  pages     = {2097-2104}
}

@article{slic,
  author  = {Achanta, Radhakrishna and Shaji, Appu and Smith, Kevin and Lucchi, Aurelien and Fua, Pascal and Süsstrunk, Sabine},
  journal = {IEEE Trans. Pattern Anal. Mach. Intell.},
  title   = {SLIC Superpixels Compared to State-of-the-Art Superpixel Methods},
  year    = {2012},
  volume  = {34},
  number  = {11},
  pages   = {2274-2282}
}

@inproceedings{lsc,
  author    = {Zhengqin Li and Jiansheng Chen},
  booktitle = {Proc. IEEE/CVF Conf. Comput. Vis. Pattern Recognit. (CVPR)},
  title     = {Superpixel segmentation using Linear Spectral Clustering},
  year      = {2015},
  volume    = {},
  number    = {},
  pages     = {1356-1363},
  doi       = {10.1109/CVPR.2015.7298741}
}

@article{superpixel_clustering_1,
  author  = {Chen, Xiaohong and Zhang, Yongshan and Feng, Xuxiang and Jiang, Xinwei and Cai, Zhihua},
  journal = {IEEE Geosci. Remote Sens. Lett.},
  title   = {Spectral-Spatial Superpixel Anchor Graph-Based Clustering for Hyperspectral Imagery},
  year    = {2023},
  volume  = {20},
  number  = {},
  pages   = {1-5},
  doi     = {10.1109/LGRS.2023.3298681}
}

@article{superpixel_clustering_2,
  author  = {Yuan, Zailong and Yang, Longshan},
  journal = {IEEE Geosci. Remote Sens. Lett.},
  title   = {Hyperspectral Image Clustering by Superpixel- Based Low-Rank Constrained Bipartite Graph Learning},
  year    = {2024},
  volume  = {21},
  number  = {},
  pages   = {1-5},
  doi     = {10.1109/LGRS.2024.3407949}
}

@ARTICLE{superpixel_clustering_3,
  author={Ding, Yao and Zhang, Zhili and Kang, Weijie and Yang, Aitao and Zhao, Junyang and Feng, Jie and Hong, Danfeng and Zheng, Qinghe},
  journal={IEEE Trans. Geosci. Remote Sens.}, 
  title={Adaptive Homophily Clustering: Structure Homophily Graph Learning With Adaptive Filter for Hyperspectral Image}, 
  year={2025},
  volume={63},
  number={},
  pages={1-13},
  doi={10.1109/TGRS.2025.3556276}}

@article{sglsc,
  author  = {Zhao, Haishi and Zhou, Fengfeng and Bruzzone, Lorenzo and Guan, Renchu and Yang, Chen},
  journal = {IEEE Trans. Geosci. Remote Sens.},
  title   = {Superpixel-Level Global and Local Similarity Graph-Based Clustering for Large Hyperspectral Images},
  year    = {2022},
  volume  = {60},
  number  = {},
  pages   = {1-16},
  doi     = {10.1109/TGRS.2021.3132683}
}

@article{ncsc,
  author  = {Cai, Yaoming and Zhang, Zijia and Ghamisi, Pedram and Ding, Yao and Liu, Xiaobo and Cai, Zhihua and Gloaguen, Richard},
  journal = {IEEE Trans. Geosci. Remote Sens.},
  title   = {Superpixel Contracted Neighborhood Contrastive Subspace Clustering Network for Hyperspectral Images},
  year    = {2022},
  volume  = {60},
  number  = {},
  pages   = {1-13}
}

@inproceedings{gcn,
  title     = {Semi-Supervised Classification with Graph Convolutional Networks},
  author    = {Thomas N. Kipf and Max Welling},
  booktitle = {Proc. Int. Conf. Learn. Represent. (ICLR)},
  year      = {2017}
}

@inproceedings{gat,
  title     = {Graph Attention Networks},
  author    = {Petar Veličković and Guillem Cucurull and Arantxa Casanova and Adriana Romero and Pietro Liò and Yoshua Bengio},
  booktitle = {Proc. Int. Conf. Learn. Represent. (ICLR)},
  year      = {2018}
}

@article{lowpass,
  author  = {Ding, Yao and Zhang, Zhili and Zhao, Xiaofeng and Cai, Yaoming and Li, Siye and Deng, Biao and Cai, Weiwei},
  journal = {IEEE Trans. Geosci. Remote Sens.},
  title   = {Self-Supervised Locality Preserving Low-Pass Graph Convolutional Embedding for Large-Scale Hyperspectral Image Clustering},
  year    = {2022},
  volume  = {60},
  number  = {},
  pages   = {1-16}
}

@ARTICLE{EGRC,
  author={Peng, Zhihao and Liu, Hui and Jia, Yuheng and Hou, Junhui},
  journal={IEEE Trans. Image Process.}, 
  title={EGRC-Net: Embedding-Induced Graph Refinement Clustering Network}, 
  year={2023},
  volume={32},
  number={},
  pages={6457-6468},
  doi={10.1109/TIP.2023.3333557}}

@ARTICLE{graph_clustering,
  author={Peng, Zhihao and Liu, Hui and Jia, Yuheng and Hou, Junhui},
  journal={IEEE Trans. Circuits Syst. Video Technol.}, 
  title={Deep Attention-Guided Graph Clustering With Dual Self-Supervision}, 
  year={2023},
  volume={33},
  number={7},
  pages={3296-3307},
  doi={10.1109/TCSVT.2022.3232604}}

@article{sdst,
  author   = {Luo, Fulin and Liu, Yi and Duan, Yule and Guo, Tan and Zhang, Lefei and Du, Bo},
  journal  = {IEEE Trans. Geosci. Remote Sens.},
  title    = {SDST: Self-Supervised Double-Structure Transformer for Hyperspectral Images Clustering},
  year     = {2024},
  volume   = {62},
  number   = {},
  pages    = {1-14},
  doi      = {10.1109/TGRS.2024.3374597}
}

@inproceedings{gae,
  author    = {Thomas N. Kipf and Max Welling},
  title     = {Variational Graph Auto-Encoders},
  booktitle = {Proc. Int. Conf. Neural Inf. Process. Syst. (NIPS)},
  year      = {2016}
}

@inproceedings{dec,
  author    = {Xie, Junyuan and Girshick, Ross and Farhadi, Ali},
  title     = {Unsupervised deep embedding for clustering analysis},
  year      = {2016},
  booktitle = {Proc. Int. Conf. Mach. Learn. (ICML)},
  pages     = {478–487}
}

@article{dgae,
  author  = {Zhang, Yongshan and Wang, Yang and Chen, Xiaohong and Jiang, Xinwei and Zhou, Yicong},
  journal = {IEEE Trans. Circuits Syst. Video Technol.},
  title   = {Spectral–Spatial Feature Extraction With Dual Graph Autoencoder for Hyperspectral Image Clustering},
  year    = {2022},
  volume  = {32},
  number  = {12},
  pages   = {8500-8511}
}

@inproceedings{moco,
  author    = {He, Kaiming and Fan, Haoqi and Wu, Yuxin and Xie, Saining and Girshick, Ross},
  booktitle = {Proc. IEEE/CVF Conf. Comput. Vis. Pattern Recognit. (CVPR)},
  title     = {Momentum Contrast for Unsupervised Visual Representation Learning},
  year      = {2020},
  volume    = {},
  number    = {},
  pages     = {9726-9735}
}

@inproceedings{byol,
  author    = {Grill, Jean-Bastien and Strub, Florian and Altch\'{e}, Florent and Tallec, Corentin and Richemond, Pierre H. and Buchatskaya, Elena and Doersch, Carl and Pires, Bernardo Avila and Guo, Zhaohan Daniel and Azar, Mohammad Gheshlaghi and Piot, Bilal and Kavukcuoglu, Koray and Munos, R\'{e}mi and Valko, Michal},
  title     = {Bootstrap your own latent: A new approach to self-supervised Learning},
  year      = {2020},
  booktitle = {Proc. Int. Conf. Neural Inf. Process. Syst. (NIPS)},
  articleno = {1786},
  numpages  = {14},
  location  = {Vancouver, BC, Canada}
}

@article{propos,
  author  = {Huang, Zhizhong and Chen, Jie and Zhang, Junping and Shan, Hongming},
  journal = {IEEE Trans. Pattern Anal. Mach. Intell.},
  title   = {Learning Representation for Clustering Via Prototype Scattering and Positive Sampling},
  year    = {2023},
  volume  = {45},
  number  = {6},
  pages   = {7509-7524}
}

@inproceedings{pscpc,
  author    = {Guan, Renxiang and Li, Zihao and Li, Xianju and Tang, Chang},
  booktitle = {Proc. IEEE Int. Conf. Acoust., Speech, Signal Process. (ICASSP)},
  title     = {Pixel-Superpixel Contrastive Learning and Pseudo-Label Correction for Hyperspectral Image Clustering},
  year      = {2024},
  volume    = {},
  number    = {},
  pages     = {6795-6799},
  doi       = {10.1109/ICASSP48485.2024.10447080}
}

@article{s2gcl,
  author  = {Yang, Aitao and Li, Min and Ding, Yao and Xiao, Xiongwu and He, Yujie},
  journal = {IEEE Trans. Geosci. Remote Sens.},
  title   = {An Efficient and Lightweight Spectral-Spatial Feature Graph Contrastive Learning Framework for Hyperspectral Image Clustering},
  year    = {2024},
  volume  = {62},
  number  = {},
  pages   = {1-14},
  doi     = {10.1109/TGRS.2024.3493096}
}

@article{spgcc,
  author  = {Qi, Jianhan and Jia, Yuheng and Liu, Hui and Hou, Junhui},
  journal = {IEEE Trans. Circuits Syst. Video Technol.},
  title   = {Superpixel Graph Contrastive Clustering With Semantic-Invariant Augmentations for Hyperspectral Images},
  year    = {2024},
  volume  = {34},
  number  = {11},
  pages   = {11360-11372},
  doi     = {10.1109/TCSVT.2024.3418610}
}

@article{survey,
  author  = {Zhai, Han and Zhang, Hongyan and Li, Pingxiang and Zhang, Liangpei},
  journal = {IEEE Geosci. Remote Sens. Mag.},
  title   = {Hyperspectral Image Clustering: Current achievements and future lines},
  year    = {2021},
  volume  = {9},
  number  = {4},
  pages   = {35-67},
  doi     = {10.1109/MGRS.2020.3032575}
}

@article{s3ulda,
  author  = {Lu, Pengyu and Jiang, Xinwei and Zhang, Yongshan and Liu, Xiaobo and Cai, Zhihua and Jiang, Junjun and Plaza, Antonio},
  journal = {IEEE Trans. Geosci. Remote Sens.},
  title   = {Spectral–Spatial and Superpixelwise Unsupervised Linear Discriminant Analysis for Feature Extraction and Classification of Hyperspectral Images},
  year    = {2023},
  volume  = {61},
  number  = {},
  pages   = {1-15}
}

@article{s2dl,
  author  = {Cui, Kangning and Li, Ruoning and Polk, Sam L and Lin, Yinyi and Zhang, Hongsheng and Murphy, James M. and Plemmons, Robert J. and Chan, Raymond H.},
  journal = {IEEE Trans. Geosci. Remote Sens.},
  title   = {Superpixel-Based and Spatially Regularized Diffusion Learning for Unsupervised Hyperspectral Image Clustering},
  year    = {2024},
  volume  = {62},
  number  = {},
  pages   = {1-18},
  doi     = {10.1109/TGRS.2024.3385202}
}

@article{sapc,
  author  = {Jiang, Guozhu and Zhang, Yongshan and Wang, Xinxin and Jiang, Xinwei and Zhang, Lefei},
  journal = {IEEE Trans. Circuits Syst. Video Technol.},
  title   = {Structured Anchor Learning for Large-Scale Hyperspectral Image Projected Clustering},
  year    = {2025},
  volume  = {35},
  number  = {3},
  pages   = {2328-2340},
  doi     = {10.1109/TCSVT.2024.3486186}
}

@article{hungarian,
  author  = {Kuhn, H. W.},
  title   = {The Hungarian method for the assignment problem},
  journal = {Nav. Res. Logist.},
  volume  = {2},
  number  = {1-2},
  pages   = {83-97},
  year    = {1955}
}

@inproceedings{bn,
  author    = {Ioffe, Sergey and Szegedy, Christian},
  title     = {Batch normalization: accelerating deep network training by reducing internal covariate shift},
  year      = {2015},
  booktitle = {Proc. Int. Conf. Mach. Learn. (ICML)},
  pages     = {448-456},
  numpages  = {9}
}

@article{infonce,
  author  = {Aaron van den Oord and Yazhe Li and Oriol Vinyals},
  title   = {Representation Learning with Contrastive Predictive Coding},
  journal = {arXiv:1807.03748},
  year    = {2019}
}

\vspace{-0.95cm}

\begin{IEEEbiography}[{\includegraphics[width=1in,height=1.25in,clip,keepaspectratio]{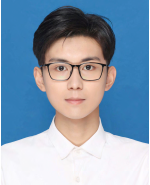}}]{Jianhan Qi}
  received the B.S. degree in Computer Science and Technology from Wuhan University of Technology, Wuhan, China, in 2022, and the M.S. degree in Software Engineering from Southeast University, Nanjing, China, in 2025. His research interests include unsupervised learning and graph neural networks.
\end{IEEEbiography}

\vspace{-0.95cm}

\begin{IEEEbiography}[{\includegraphics[width=1in,height=1.25in,clip,keepaspectratio]{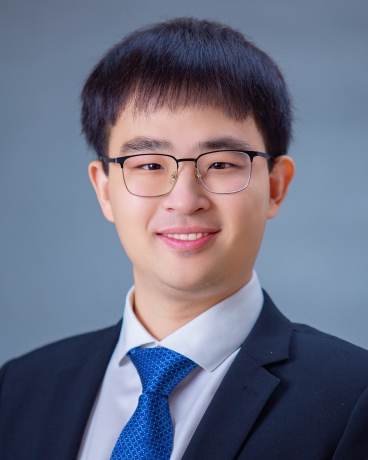}}]{Yuheng Jia}
  (Member, IEEE) received the B.S. degree in automation and the M.S. degree in control theory and engineering from Zhengzhou University, Zhengzhou, China, in 2012 and 2015, respectively, and the Ph.D. degree in computer science from the City University of Hong Kong, Hong Kong, China, in 2019. He is currently an Associate Professor with the School of Computer Science and Engineering, Southeast University, Nanjing, China. His research interests include machine learning and data representation, such as un/semi/weakly-supervised learning, high-dimensional data modeling and analysis, and low-rank tensor/matrix approximation and factorization.
\end{IEEEbiography}

\vspace{-0.95cm}

\begin{IEEEbiography}[{\includegraphics[width=1in,height=1.25in,clip,keepaspectratio]{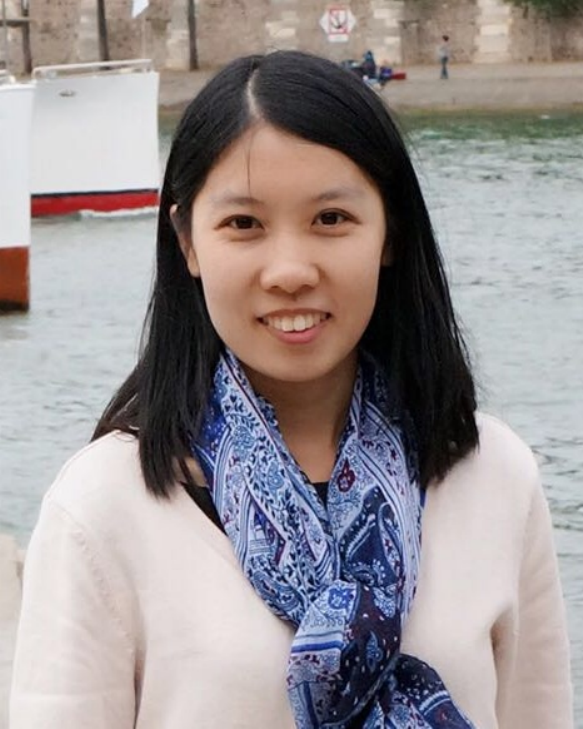}}]{Hui Liu}
  received the B.Sc. degree in communication engineering from Central South University, Changsha, China, the M.Eng. degree in computer science from Nanyang Technological University, Singapore, and the Ph.D. degree in computer science from the City University of Hong Kong, Hong Kong, China. From 2014 to 2017, she was a Research Associate with the Maritime Institute, Nanyang Technological University. She is currently an Assistant Professor with the School of Computing Information Sciences, Caritas Institute of Higher Education, Hong Kong, China. Her current research interests include image processing and machine learning.
\end{IEEEbiography}

\vspace{-0.95cm}

\begin{IEEEbiography}[{\includegraphics[width=1in,height=1.25in,clip,keepaspectratio]{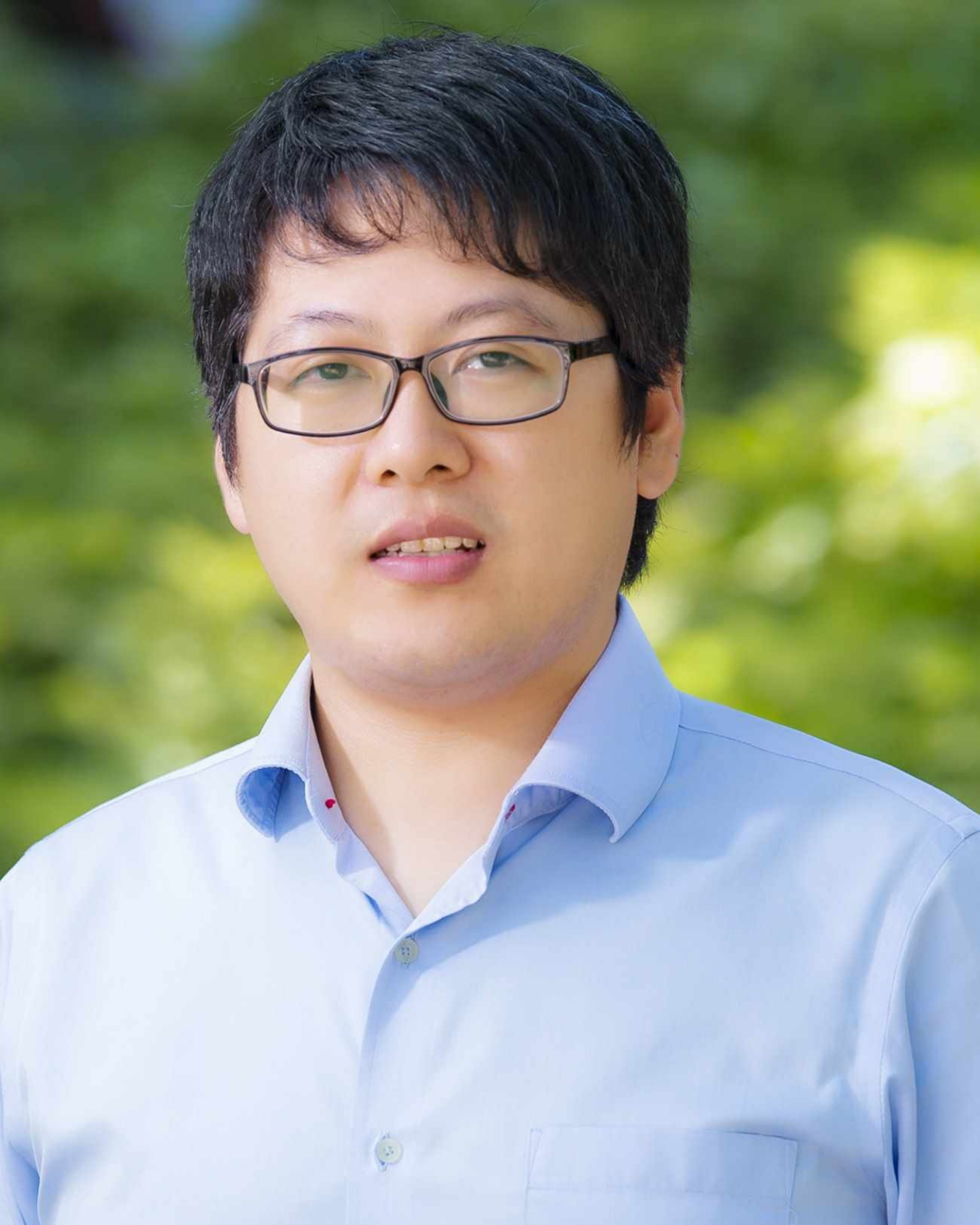}}]{Junhui Hou}
  (Senior Member, IEEE) is an Associate Professor with the Department of Computer Science, City University of Hong Kong. He holds a B.Eng. degree in information engineering (Talented Students Program) from the South China University of Technology, Guangzhou, China (2009), an M.Eng. degree in signal and information processing from Northwestern Polytechnical University, Xi’an, China (2012), and a Ph.D. degree from the School of Electrical and Electronic Engineering, Nanyang Technological University, Singapore (2016). His research interests are multi-dimensional visual computing. Dr. Hou received the Early Career Award (3/381) from the Hong Kong Research Grants Council in 2018 and the NSFC Excellent Young Scientists Fund in 2024. He has served or is serving as an Associate Editor for IEEE Transactions on Visualization and Computer Graphics, IEEE Transactions on Image Processing, IEEE Transactions on Multimedia, and IEEE Transactions on Circuits and Systems for Video Technology.
\end{IEEEbiography}

\vfill
\end{document}